\documentclass[singlecolumn]{fairmeta}

\usepackage[utf8]{inputenc} 
\usepackage[T1]{fontenc}    
\usepackage{hyperref}       
\usepackage{url}            
\usepackage{booktabs}       
\usepackage{amsfonts}       
\usepackage{nicefrac}       
\usepackage{microtype}      
\usepackage{xcolor}         

\usepackage{latexsym}
\usepackage{colortbl}
\usepackage{multirow}
\usepackage{amsmath}
\usepackage{graphicx}		
\usepackage{booktabs}
\usepackage{algorithm}
\usepackage{algpseudocode}
\usepackage{longtable}
\usepackage{amssymb}
\usepackage{xspace}
\usepackage{tabularx} 
\usepackage{enumitem}
\usepackage{amssymb}
\usepackage{bbm}
\usepackage[skip=6pt]{caption}
\usepackage{float}
\usepackage{wrapfig}
\usepackage{cleveref}
\usepackage{amsthm}

\usepackage{amsmath}
\usepackage{amssymb}
\usepackage{mathtools}
\usepackage{bm}
\usepackage{algorithm}
\usepackage{algpseudocode}
\usepackage{paralist}
\usepackage{tabularx}
\usepackage[most]{tcolorbox}
\usepackage[dvipsnames,table,xcdraw]{xcolor}
\usepackage{multirow}
\usepackage{tikz}
\usepackage{pgfplots}
\usepackage{wrapfig}
\usetikzlibrary{pgfplots.groupplots, matrix}
\pgfplotsset{compat=1.18}
\usepackage{enumitem}
\usepackage{comment}
\usepackage{graphicx}
\usepackage{array}
\usepackage{multirow}
\usepackage{stackengine}

\definecolor{purple}{HTML}{c994c7}
\definecolor{navyblue}{HTML}{75207B}   
\definecolor{citecolor}{HTML}{75207B}  
\definecolor{lightgray}{gray}{0.9}
\definecolor{blanchedalmond}{rgb}{1.0, 0.92, 0.8}
\definecolor{cerise}{rgb}{0.871, 0.192, 0.388}

\definecolor{TaskBG}{HTML}{EFE6FF}        
\definecolor{StateBG}{HTML}{F5F5F7}       
\definecolor{ExpertBG}{HTML}{EAF7EA}      
\definecolor{IWMBG}{HTML}{FDECF3}         
\definecolor{SRBG}{HTML}{E6F2FF}          

\newtcolorbox{promptbox}[2][]{%
  enhanced, breakable, colframe=black, colback=white, colbacktitle=black!75,
  boxrule=0.8pt, arc=2pt, left=3pt, right=3pt, top=2pt, bottom=2pt,
  title={#2}, fonttitle=\bfseries, coltitle=white, #1}

 \usepackage{subcaption}

\newcommand{\secref}[1]{\S\ref{#1}}

\newcommand{\se}{\mathcal{R}}

\renewcommand{\secref}[1]{\S\ref{#1}}

\newcommand{\alg}{\textsc{ParSer}\xspace}

\title{{P\scalebox{0.75}{AR}S\scalebox{0.75}{ER}\xspace}: Read in Parallel, Reason in Depth for Long-Context LLM Agents}

\author[*]{Kun Li}
\author[*]{Zexuan Qiu}
\author[*]{Tianhua Zhang}
\author{Irwin King}
\author{Helen Meng}
\affiliation{The Chinese University of Hong Kong}
\contribution[*]{Equal contribution.}
\metadata[Email]{\texttt{\{li.kun, qzexuan, thzhang\}@link.cuhk.edu.hk}}
\metadata[Project Page]{\url{https://cuhk-parser.github.io/}}

\abstract{
Reasoning over documents far beyond an LLM's context window remains challenging, as evidence may be sparsely distributed across hundreds of thousands of tokens. Sequential memory agents stream documents chunk by chunk into a compact recurrent memory, enabling bounded-context processing of arbitrarily long inputs. However, because every reading step immediately updates the memory on which subsequent processing depends, these agents couple document traversal with sequential reasoning. This coupling makes the reasoning sensitive to evidence position and forces the sequential inference path to grow with document length.
We introduce \alg (\textbf{Pa}rallel \textbf{R}eading, \textbf{Se}quential \textbf{R}easoning), a framework that separates document reading from question reasoning. To realize this separation, \alg assigns local reading to chunk-bound subagents and global reasoning to a lead agent. The subagents read in parallel under the lead agent's queries; the lead agent integrates their findings and iteratively refines its queries as evidence accumulates. This decoupled design concentrates all learnable behavior in the lead agent and optimizes it with reinforcement learning, while the lightweight chunk readers remain frozen off-the-shelf models.
On multi-hop QA with contexts ranging from 7K to 896K tokens, \alg with a 4B backbone outperforms the strongest sequential memory baseline by 5.7 points on average and by 12.0 points at 896K tokens. With a 9B backbone, \alg surpasses DeepSeek-V4-Pro by 6.3 points. Experiments show that \alg remains robust to changes in evidence position, order, and distance that cause large accuracy swings in sequential methods. By parallelizing document coverage, \alg reduces inference latency by up to $11\times$.
}

\begin{document}

\maketitle

\section{Introduction}
\label{sec:intro}

Reasoning over long documents is a core capability for large language models, yet remains challenging: in tasks such as multi-document QA or legal analysis, the evidence for a single question can be scattered across hundreds of thousands of tokens. Despite context windows now reaching a million tokens or more \citep{deepseekai2026deepseekv4}, model accuracy degrades as context grows, a phenomenon known as \emph{context rot} \citep{hong2025context}. One well-documented manifestation is positional bias: evidence placed away from the boundaries of the input is systematically ignored \citep{liu-etal-2024-lost}. We observe the same on multi-hop QA, where direct full-context answering drops by tens of points as documents lengthen from $7$K to $896$K tokens, even for million-token models. Since accuracy degrades even when the window is far from full, the bottleneck is no longer capacity; the open question is how to select what to read and in what order to reason about it.

One response to the selection problem is the sequential memory paradigm: the document is read chunk by chunk, and at each step the agent compresses the current chunk together with its previous memory into an updated memory. The final answer is generated from this memory alone. MemAgent \citep{yu2026memagent} and its follow-ups \citep{shi2026look, sheng2026grumem} train this recurrent workflow end-to-end with reinforcement learning, handling documents of millions of tokens within a small context window. These methods address a genuinely hard problem: bounded working memory, arbitrary input length, and end-to-end trainability.

However, these memory-based methods share a structural commitment: \emph{the document is traversed exactly once, in order}. This single-pass recurrent structure imposes two constraints: evidence must be evaluated through a repeatedly compressed prefix state, and every chunk update depends on the preceding one. The first leads to a \textbf{sensitivity to evidence placement}: because each chunk is compressed before the rest of the document has been seen, the agent judges relevance under a strict information deficit. In particular, its accuracy could be affected by the absolute position of evidence, the logical order among evidence pieces, and the distance between them. This sensitivity is most damaging in multi-hop reasoning, where the answer depends on scattered evidence whose relevance emerges only incrementally. The second leads to \textbf{inference latency}: since each chunk update depends on the output of the previous one, the $T$ steps form an irreducibly sequential chain whose wall-clock cost grows linearly with document length, regardless of available parallelism. Subsequent work in this sequential memory paradigm has largely been a sequence of patches to these two symptoms, often trading one for the other. \citet{shi2026look} adds a callback module that retrieves earlier memory states to counter position bias, at the price of extra retrieval on the sequential path; \citet{sheng2026grumem} adds gates that skip evidence-free chunks to save computation, but the sequential chain remains intact because the agent must still scan up to the last required evidence.

\emph{The order in which a long document is read is imposed by the document; the order in which a question is reasoned about is imposed by the question.} Sequential memory agents let the first drive the second. We introduce \alg (\textbf{Pa}rallel \textbf{R}eading, \textbf{Se}quential \textbf{R}easoning), which decouples the two orders. Rather than tying sequential depth to document length, \alg reads all chunks in parallel and reserves sequential computation only for the reasoning the question demands. To achieve this decoupling, \alg assigns reading and reasoning to two separate roles. A \textbf{lead agent}, blind to the document, \textit{reasons} about the question in a ReAct-style \citep{yao2023react} loop of interleaved thinking and action. A bank of \textbf{subagents} (one per chunk) \textit{read} only their own chunk and extract evidence in parallel for a given query. At each round, the lead agent \emph{scatters} queries to all subagents, then \emph{gathers} their findings and decides what to ask next, iterating until committing to an answer.

This scatter--gather architecture changes the dependency structure of long-document processing. Sequential memory requires $T$ dependent updates in sequence; \alg replaces this with $K$ scatter--gather rounds where all chunk-level calls execute in parallel, so latency scales with reasoning depth rather than document length. Furthermore, every chunk is re-read under a freshly formulated query at each round, eliminating the chunk-level position bias inherent in sequential traversal. Because no chunk is permanently discarded, the lead agent can condition each new query on previously discovered evidence,  which is essential for multi-hop reasoning where a single static query cannot identify downstream hops \citep{zhou2024llmxmapreduce, zhao2024longagent, xu2026divideconquer}. Although this requires reading every chunk over multiple rounds, cross-round KV-cache reuse avoids repeated prefill computation and sparse subagent outputs limit decoding computation. Finally, this decoupled design simplifies training: we train \emph{only} the lead agent with Reinforcement Learning with Verifiable Reward \citep{shao2024deepseekmathpushinglimitsmathematical, deepseekai2025deepseekr1incentivizingreasoningcapability}. Since each subagent's task is simple (locate evidence for a focused query in a short span), an off-the-shelf model suffices.

We evaluate \alg on multi-hop long-context question answering, including the in-distribution HotpotQA \citep{yang2018hotpotqa} and the out-of-distribution 2WikiMultiHopQA \citep{xanh2020_2wikimultihop}, with context ranging from $7$K to $896$K tokens. On HotpotQA, \alg with 4B and 9B backbones achieves average accuracies of $84.6\%$ and $86.8\%$ respectively, outperforming the strongest sequential memory baseline by $5.7$ and $6.7$ percentage points; at the longest setting ($896$K tokens) the gaps widen to $12.0$ and $9.9$ percentage points, as sequential methods degrade sharply with length while \alg remains stable. Scaling to a 9B backbone, \alg achieves an average of $86.8\%$, surpassing DeepSeek-V4-Pro \citep{deepseekai2026deepseekv4}, which natively supports a one-million-token context, by $6.3$ percentage points.
Controlled experiments that independently perturb the absolute position, the logical order, and the relative distance of evidence within context confirm the source of this stability: sequential memory agents exhibit large accuracy swings as any of these factors changes, whereas \alg remains nearly flat across all three conditions. On inference latency, parallel reading yields an $11\times$ reduction at $896$K tokens under single concurrency ($876$s vs.\ $78$s per sample relative to MemAgent) and maintains a $1.7\times$ advantage under a concurrency of $16$ ($102$s vs.\ $59$s). 

\section{Method}
\label{sec:method}
\subsection{Problem Formulation}

For the task of long-context question answering (QA), an agent is required to give an answer $\bm{a}$ to the question $\bm{q}$, conditioned on a corresponding long document $\bm{D}$. The document $\bm{D}$ can be extremely long, such as hundreds of thousands of tokens or even more. Typically, due to the limited LLM context window, $\bm{D}$ is split into a set of fixed-size chunks $\{\bm{d}_1, \bm{d}_2, \ldots, \bm{d}_{T}\}$ for processing. To answer the question $\bm{q}$, the agent needs to accurately locate and then reason over a few pieces of key evidence,  which are sparsely distributed within $\bm{D}$.

As illustrated in the upper panel of Figure~\ref{fig:workflow}, sequential memory methods (e.g., MemAgent, ReMemR1) formulate long-context reasoning as a sequential, recurrent, and chunk-by-chunk process: throughout the entire reasoning process, the agent maintains a textual memory, which stores key summaries of the chunks. The update operation relies on the question, the previous memory, and the current chunk to produce the updated memory. Consequently, the update operations for chunk $\bm{d}_t$ with $0<t\leq T$ must execute sequentially. 


\subsection{Workflow: Parallel Reading, Sequential Reasoning}
\label{sec:workflow}

\begin{figure*}[t]
\centering
\includegraphics[width=0.82\textwidth]{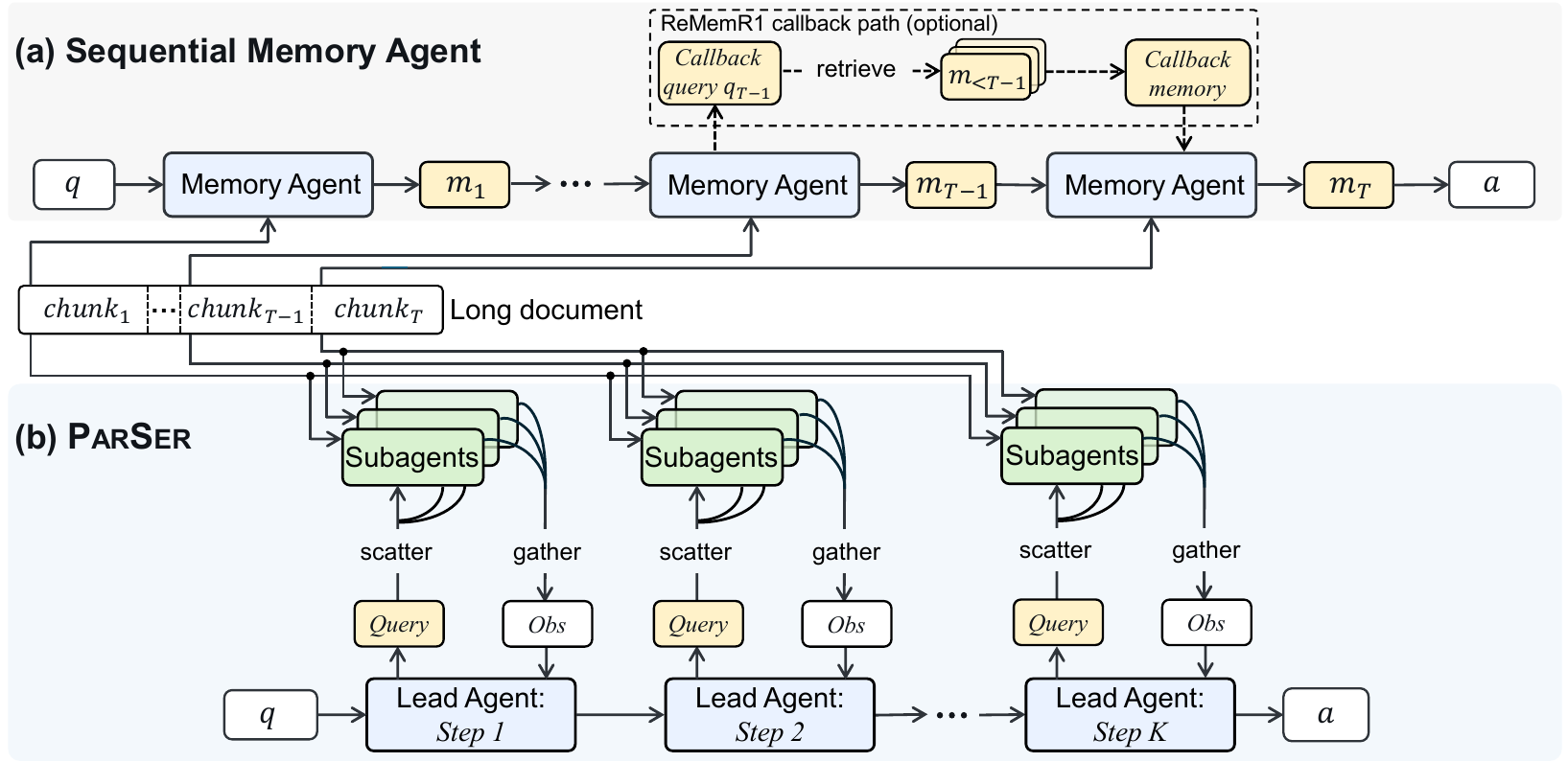}
\caption{Comparison between \alg and sequential memory methods. (a) Sequential memory traverses $T$ chunks through a chain of dependent reading steps. (b) \alg turns the chunks into parallel width: at each round, all chunks are read concurrently under the same query, while reasoning proceeds across rounds.}
\label{fig:workflow}
\end{figure*}

Sequential memory turns document length into dependency depth: processing $\bm{d}_t$ cannot start until the memory from $\bm{d}_{<t}$ has been written. \alg instead turns document length into \emph{parallel width}. It assigns one subagent to each chunk and organizes their interaction with a lead agent through repeated \textbf{Scatter-Gather} rounds (the lower panel of Figure~\ref{fig:workflow}; Algorithm~\ref{alg:longmas}). At round $k$, the lead agent \emph{scatters} one or more focused queries across all $T$ subagents; the subagents inspect their respective chunks concurrently, and their local findings are \emph{gathered} as the observation $\se$ for the lead agent. Each round therefore covers the entire document in parallel: increasing $T$ adds parallel readers rather than dependent reading steps.

\textbf{Parallel chunk readers.} Each subagent is bound to one chunk and, given a lead-agent query, returns a finding grounded only in that chunk. Chunks can be irrelevant to a given focused query, so subagents may abstain from answering; abstentions are dropped during gathering. Appendix~\ref{app:subagent-prompt} shows the subagent prompt.
All $T$ subagents run in parallel at every round, each receiving the same query while retaining its fixed chunk assignment.
Binding each subagent to a short chunk also keeps its effective context compact, mitigating the context rot issue that arises as input length grows. Together with the decomposed query, this makes the subagent's reading task simpler; we thus let the subagents run in non-thinking mode. We deploy the subagents with SGLang \citep{zheng2024sglang} and achieve parallel execution across all subagents through concurrent request dispatching. To reduce repeated chunk prefill across rounds, prompts with fixed instruction-chunk prefixes and cache-aware request routing  maximize cross-round KV-cache reuse for subagent inference (Appendix~\ref{app:training_setup}).

\textbf{Question-driven reasoner.} The lead agent controls this parallel reading. It takes as input $\bm{q}$, but never $\bm{D}$ or any chunk $\bm{d}_t$, so it reasons about the question rather than the document. Appendix~\ref{app:lead-prompt} gives the lead-agent prompt. The lead agent conducts this reasoning in a multi-step ReAct \citep{yao2023react} loop of interleaved thinking and action---after thinking, it either performs a scatter-gather operation or commits to a final answer. Queries in the latest round are conditioned on the reasoning history including previously gathered findings. Only this reasoning process is sequential; document-wide reading remains parallel in every round. 
The loop terminates when the lead agent answers or reaches the maximum number of rounds. The number of rounds actually executed, $K$, is determined by the reasoning hops required by $\bm{q}$ rather than the number of chunks $T$. 

Parallel reading yields two immediate consequences. 
First, every chunk is inspected under the same query in each round and can be revisited under a newly formulated query. Access to evidence is therefore symmetric with respect to chunk position. 
Second, parallel reading removes document coverage from the sequential critical path: the dependent reasoning depth is $K$ rounds rather than $T$ chunks, and long documents satisfy $T \gg K$, leading to lower wall-clock latency.
On the other hand, parallelizing the readers raises two natural concerns: whether independent chunk reading undermines cross-chunk dependencies required for answering complex questions, and whether repeatedly reading all chunks incurs excessive computation.

\textbf{Adaptive parallel reading across rounds.} Although processing chunks independently prevents each subagent from observing dependencies that span multiple chunks, \alg does not ask subagents to solve the original multi-hop question. Instead, dependencies that span chunks are carried across reasoning rounds by lead agent: the lead agent decomposes a complex question into focused queries that can typically be addressed independently within individual chunks; findings gathered at round $k$ condition the query at round $k+1$. Cross-chunk dependencies are thus resolved through successive parallel reading rounds. \alg relocates such composition from document-ordered memory propagation to the question-driven query chain.

\textbf{Efficiency through sparsity.} The same decomposition also makes subagent communication sparse. For each query, only a small number of subagents return findings, while most emit only a short abstention and their responses are dropped. Sequential memory methods, in contrast, generate a memory update (with hundreds or thousands of tokens) after every chunk. While multi-round reading may preserve or increase prefill computation (depending on whether KV-cache reuse is available), \alg reduces decoding computation through substantially fewer generated tokens. Appendix~\ref{app:compute_analysis} gives detailed analysis.

Taken together, \alg separates query-conditioned local reading from evidence-conditioned global reasoning, composing cross-chunk evidence through successive parallel reading rounds.

\subsection{Optimization: Agentic Reinforcement Learning}

The decoupling of reading from reasoning also determines what we train. For the comparatively simple reading task, a frozen off-the-shelf model suffices as the subagent. What still has to be learned is the lead agent's policy: how to determine the next action based on reasoning history. We therefore train \emph{only} the lead agent $\pi_{\theta}$ and keep the subagents frozen.

We optimize the lead agent with Reinforcement Learning with Verifiable Reward (RLVR; \citealt{deepseekai2025deepseekr1incentivizingreasoningcapability}). The reward is a binary exact-match score $r(\bm{q}, \bm{y})=\texttt{EM}(\bm{a}_{pred},\bm{a}_{gold})$, where $\bm{a}_{pred}$ is the answer extracted from the reasoning trajectory $\bm{y}$ and $\bm{a}_{gold}$ is the ground truth. We do not use format rewards, as the lead agent uses the backbone's native multi-turn tool-calling format.
With this reward, we use Group Relative Policy Optimization (GRPO; \citealt{shao2024deepseekmathpushinglimitsmathematical}) to maximize
\begin{equation}\label{eq:grpo}
\scalebox{0.78}{$\displaystyle
\mathcal{J}_{\mathrm{GRPO}}(\theta)
= \mathbb{E}_{\bm{q}, \{ \bm{y}_i \}_{i=1}^{G} \sim \pi_{\text{old}}(\cdot \mid \bm{q}; \mathcal{S}(\bm{D}))}
\Biggl[
\frac{1}{G} \sum_{i=1}^{G} \frac{1}{|\bm{y}_i|} \sum_{t=1}^{|\bm{y}_i|}
\min \bigl( \rho_{i,t}(\theta) \hat{A}_{i,t},\;
\operatorname{clip}(\rho_{i,t}(\theta), 1{-}\epsilon, 1{+}\epsilon)\, \hat{A}_{i,t} \bigr)
- \beta \mathbb{D}_{\mathrm{KL}}[\pi_{\theta} \| \pi_{\text{ref}}]
\Biggr]
$}
\end{equation}
where $\rho_{i,t}(\theta)=\frac{\pi_{\theta}(\bm{y}_{i,t} \mid \bm{q}, \bm{y}_{i,<t}; \mathcal{S}(\bm{D}))}{\pi_{\text{old}}(\bm{y}_{i,t} \mid \bm{q}, \bm{y}_{i,<t}; \mathcal{S}(\bm{D}))}$ is the importance ratio, $\mathcal{S}(\bm{D})$ denotes the frozen subagents bound to the chunks of $\bm{D}$, $\epsilon$ is the PPO clipping hyperparameter, $\beta$ is the KL regularization coefficient, and $\hat{A}_{i,t}$ denotes the advantage computed from the relative rewards of outputs in each group.
We mask observation tokens, the findings gathered from $\mathcal{S}(\bm{D})$, so the policy gradient is applied only to tokens generated by the lead agent.

\section{Experiments}
\label{sec:exp}
\subsection{Implementation}
Following prior work \citep{yu2026memagent, shi2026look}, we utilize multi-hop long-context question answering tasks for our training. We synthesized $32{,}768$ training samples from the HotpotQA \citep{yang2018hotpotqa} dataset by following the recipe of \cite{yu2026memagent}, and each synthetic sample has a context composed of 200 paragraphs, with a total token length of $\sim28$K tokens. More details of sample synthesis are in Appendix~\ref{app:data_details}. 

We choose Qwen3.5-4B and Qwen3.5-9B \citep{qwen35blog} as backbone models. During training, we impose an upper bound of 9 total turns, i.e., $K \leq 9$; each turn is capped at 2048 tokens.
The document is chunked into at most 512 tokens per chunk. The subagents use a temperature of 0.7 and a 512-token generation budget. To streamline the aggregation of findings from subagents, we instruct the subagents to structure their output in JSON format. At inference, we increase the cap of $K$ to 12 and the chunk size to $4{,}096$ tokens.\footnote{To pursue faster training, we intentionally select a smaller chunk size for the training phase, despite the resulting mismatch with inference-time chunk sizes. As given in Equation~\eqref{eq:prefill_complexity}, the prefill cost $\mathcal{O}(n^2/T)$ decreases as the number of chunks $T$ grows.}

We optimize the lead agent using a learning rate of $1e-6$, a mini-batch size of 128, and 15 warm-up steps. We apply a PPO clipping with $\epsilon = 0.2$, and KL regularization with $\beta = 1e-3$. The group size of rollouts is set to 5.
We train \alg on top of VERL \citep{10.1145/3689031.3696075} framework with Megatron backend training and SGLang \citep{zheng2024sglang} rollout service. To avoid leaving the subagents idle during policy updates, we adopt fully asynchronous RL: all training runs on 6 H100 GPUs; 4 GPUs serve rollouts and 2 GPUs train the actor. We use 10 additional H100 GPUs to deploy subagents with SGLang. We let Qwen3.5-4B serve as the subagent for both 4B and 9B lead agents. More details of training and inference are in Appendix~\ref{app:training_setup} and \ref{app:inference_details}, respectively.

\subsection{Baselines \& Evaluation}
\label{sec:baselines}
We compare our method against the following baselines: (1) \textbf{Long-context LLMs}, including Qwen3.5 \citep{qwen35blog} and DeepSeek-V4-Pro \citep{deepseekai2026deepseekv4}, which take the question together with the entire associated document as input to generate direct answers. DeepSeek-V4-Pro (preview, 2026-04-24) supports 1M-token context, and we set its reasoning effort mode as \texttt{Max}, the highest level of reasoning effort of this model. For Qwen3.5, we use the models' default configuration without context extension for evaluations within 262K tokens; furthermore, to support evaluations beyond the models' native 262K-token context window, we apply YaRN \citep{peng2024yarn}-based RoPE scaling with a factor of 4.0, extending its context window to approximately 1M tokens. The temperature for them is set to $0$.
(2) \textbf{Agentic RAG} \citep{jin2025searchr}, in which an agent iteratively searches relevant chunks from the long document through Qwen3-Embedding-4B\citep{qwen3embedding} and makes reasoning.
(3) \textbf{Sequential memory agents}, such as MemAgent \citep{yu2026memagent} and ReMemR1 \citep{shi2026look}. The last two lines of baselines are trained through RL, and we reimplement them under configurations identical to our approach, covering training data and backbone models. Appendix~\ref{app:baseline_implementation} shows the details of the re-implementation.

For evaluation, we use the in-distribution HotpotQA \citep{yang2018hotpotqa} and the out-of-distribution 2WikiMultiHopQA \citep{xanh2020_2wikimultihop}. We use the HotpotQA test samples released by \citet{yu2026memagent}, and regenerate the 2WikiMultiHopQA ones with \citeauthor{shi2026look}'s \citeyearpar{shi2026look} public script. To enable comprehensive evaluations across diverse document lengths, the documents of the test samples have lengths ranging from $7$K to $896$K tokens.
Following \cite{yu2026memagent} and \cite{shi2026look}, we report \texttt{Sub\_EM} as the evaluation metric. For each training method, we select the checkpoint that achieves the best in-distribution overall performance and report the average score over 3 runs.
We also evaluate \alg on ten non-QA tasks from RULER~\citep{hsieh2024ruler} in Appendix~\ref{app:ruler_ood}.

\subsection{Main Results}
\label{sec:main_results}
\begin{table}[t]
    \centering
    \caption{Long-context QA results on HotpotQA~\citep{yang2018hotpotqa} and 2WikiMultiHopQA~\citep{xanh2020_2wikimultihop}. Values are accuracy (Sub\_EM, \%). Within each Qwen backbone block, the best result in each column is \textbf{bolded}.}
    \label{tab:main_hqa}
    \label{tab:main_2wiki}
    \scriptsize
    \resizebox{\linewidth}{!}{%
    \begin{tabular}{llccccccccc}
        \multicolumn{11}{c}{(a) Accuracy on HotpotQA (In-Distribution)} \\
        \midrule
        \multirow{2}{*}{\textbf{Backbone}} & \multirow{2}{*}{\textbf{Method}} & \multicolumn{8}{c}{\# \textbf{Paragraphs (Total Length)}} & \multirow{2}{*}{\textbf{Avg.}} \\
        \cmidrule(lr){3-10}
         & & \shortstack{50\\(7K)} & \shortstack{100\\(14K)} & \shortstack{200\\(28K)} & \shortstack{400\\(56K)} & \shortstack{800\\(112K)} & \shortstack{1600\\(224K)} & \shortstack{3200\\(448K)} & \shortstack{6400\\(896K)} & \\
        \midrule
        \multirow{2}{*}{DeepSeek-V4-Pro}
            & Full-context (non-think) & 78.12 & 77.34 & 76.56 & 79.69 & 77.34 & 75.78 & 73.44 & 62.50 & 75.10 \\
            & Full-context (think-max) & 82.03 & 82.81 & 80.47 & 80.47 & 80.47 & 81.25 & 77.34 & 78.91 & 80.47 \\
        \midrule
        \multirow{6}{*}{Qwen3.5-4B}
            & Full-context (non-think) & 75.78 & 75.78 & 71.88 & 74.22 & 67.97 & 61.72 & 53.13 & 34.38 & 64.36 \\
            & Full-context (think) & 80.47 & 78.91 & 78.12 & 76.56 & 74.22 & 60.94 & 46.09 & 31.25 & 65.82 \\
            & Agentic RAG & - & - & - & - & - & - & - & - & - \\
            & MemAgent & 81.25 & 81.51 & 83.59 & 77.86 & 77.86 & 79.69 & 73.96 & 72.92 & 78.58 \\
            & ReMemR1 & 82.03 & 79.95 & 82.03 & 78.91 & 77.86 & 79.95 & 77.08 & 73.44 & 78.91 \\
            & \cellcolor{metabg}\alg & \cellcolor{metabg}\textbf{85.68} & \cellcolor{metabg}\textbf{84.64} & \cellcolor{metabg}\textbf{83.60} & \cellcolor{metabg}\textbf{85.68} & \cellcolor{metabg}\textbf{85.42} & \cellcolor{metabg}\textbf{83.07} & \cellcolor{metabg}\textbf{83.07} & \cellcolor{metabg}\textbf{85.42} & \cellcolor{metabg}\textbf{84.57} \\
        \midrule
        \multirow{6}{*}{Qwen3.5-9B}
            & Full-context (non-think) & 75.00 & 72.66 & 72.66 & 71.88 & 70.31 & 65.62 & 58.59 & 47.66 & 66.80 \\
            & Full-context (think) & 77.34 & 74.22 & 78.91 & 76.56 & 77.34 & 65.62 & 53.91 & 46.88 & 68.85 \\
            & Agentic RAG & - & - & - & - & - & - & - & - & - \\
            & MemAgent & 81.77 & 80.73 & 82.03 & 79.95 & 79.95 & 81.25 & 79.69 & 75.00 & 80.05 \\
            & ReMemR1 & 81.25 & 79.69 & 77.60 & 78.39 & 78.13 & 78.13 & 77.86 & 76.04 & 78.39 \\
            & \cellcolor{metabg}\alg & \cellcolor{metabg}\textbf{86.72} & \cellcolor{metabg}\textbf{88.80} & \cellcolor{metabg}\textbf{87.24} & \cellcolor{metabg}\textbf{85.68} & \cellcolor{metabg}\textbf{86.46} & \cellcolor{metabg}\textbf{86.72} & \cellcolor{metabg}\textbf{86.72} & \cellcolor{metabg}\textbf{85.94} & \cellcolor{metabg}\textbf{86.79} \\
        \midrule
        \multicolumn{11}{c}{\shortstack{\strut\\(b) Accuracy on 2WikiMultiHopQA (Out-of-Distribution)}} \\
        \midrule
        \multirow{2}{*}{\textbf{Backbone}} & \multirow{2}{*}{\textbf{Method}} & \multicolumn{8}{c}{\# \textbf{Paragraphs (Total Length)}} & \multirow{2}{*}{\textbf{Avg.}} \\
        \cmidrule(lr){3-10}
         & & \shortstack{50\\(7K)} & \shortstack{100\\(14K)} & \shortstack{200\\(28K)} & \shortstack{400\\(56K)} & \shortstack{800\\(112K)} & \shortstack{1600\\(224K)} & \shortstack{3200\\(448K)} & \shortstack{6400\\(896K)} & \\
        \midrule
        \multirow{2}{*}{DeepSeek-V4-Pro}
            & Full-context (non-think) & 89.84 & 86.72 & 87.50 & 87.50 & 87.50 & 77.34 & 78.91 & 68.75 & 83.01 \\
            & Full-context (think-max) & 90.62 & 89.06 & 92.19 & 90.62 & 92.19 & 87.50 & 82.81 & 80.47 & 88.18 \\
        \midrule
        \multirow{6}{*}{Qwen3.5-4B}
            & Full-context (non-think) & 85.94 & 85.94 & 82.81 & 78.12 & 74.22 & 69.53 & 66.41 & 46.88 & 73.73 \\
            & Full-context (think) & 88.28 & \textbf{88.28} & \textbf{88.28} & 83.59 & 82.81 & 70.31 & 63.28 & 39.06 & 75.49 \\
            & Agentic RAG & - & - & - & - & - & - & - & - & - \\
            & MemAgent & 67.45 & 73.18 & 69.01 & 61.46 & 54.17 & 54.43 & 60.16 & 45.05 & 60.61 \\
            & ReMemR1 & \textbf{88.54} & 86.85 & 83.20 & 83.07 & 76.83 & 67.71 & 72.40 & 60.68 & 77.41 \\
            & \cellcolor{metabg}\alg & \cellcolor{metabg}86.98 & \cellcolor{metabg}84.64 & \cellcolor{metabg}\textbf{88.28} & \cellcolor{metabg}\textbf{87.76} & \cellcolor{metabg}\textbf{87.50} & \cellcolor{metabg}\textbf{85.94} & \cellcolor{metabg}\textbf{88.28} & \cellcolor{metabg}\textbf{86.98} & \cellcolor{metabg}\textbf{87.04} \\
        \midrule
        \multirow{6}{*}{Qwen3.5-9B}
            & Full-context (non-think) & 80.47 & 82.81 & 81.25 & 80.47 & 74.22 & 78.91 & 61.72 & 50.00 & 73.73 \\
            & Full-context (think) & \textbf{88.28} & \textbf{89.06} & \textbf{92.19} & 85.16 & \textbf{89.06} & 82.81 & 57.81 & 52.34 & 79.59 \\
            & Agentic RAG & - & - & - & - & - & - & - & - & - \\
            & MemAgent & 80.73 & 82.03 & 78.91 & 75.26 & 70.57 & 65.63 & 75.78 & 60.94 & 73.73 \\
            & ReMemR1 & 84.12 & 87.50 & 77.87 & 83.59 & 76.30 & 74.74 & 79.43 & 70.57 & 79.27 \\
            & \cellcolor{metabg}\alg & \cellcolor{metabg}87.76 & \cellcolor{metabg}88.80 & \cellcolor{metabg}88.80 & \cellcolor{metabg}\textbf{89.06} & \cellcolor{metabg}87.24 & \cellcolor{metabg}\textbf{89.58} & \cellcolor{metabg}\textbf{88.54} & \cellcolor{metabg}\textbf{88.02} & \cellcolor{metabg}\textbf{88.48} \\
        \bottomrule
    \end{tabular}%
    }
\end{table}

Table~\ref{tab:main_hqa} presents all methods' results on two benchmarks. \alg consistently demonstrates higher accuracy than all baselines across all subsets on HotpotQA and long-document subsets (with $\geq1600$ paragraphs) on 2WikiMultiHopQA. For the three long-context LLMs, the baselines that incorporate full documents as input suffer severe performance degradation as the document length increases. While the sequential memory agent methods can alleviate this issue to some extent by storing salient information in a memory buffer, \alg exhibits minimal or even zero performance drop on both benchmarks. 

Among the training-based methods, MemAgent and ReMemR1 perform far worse under out-of-distribution conditions than under in-distribution scenarios; in contrast, \alg maintains favorable performance on out-of-distribution cases. 
We attribute this robustness to our decoupling of reading from reasoning. The lead agent never sees the document, so its training develops general reasoning capabilities rather than learning to generate document-specific memorization, as memory-based baselines do. Meanwhile, the subagents, with access to the document, remain frozen as off-the-shelf models. This decoupled reading-and-reasoning strategy thereby helps mitigate overfitting to the training documents.
\section{Analysis}
\label{sec:analysis}
We compare \alg with and without RL in Appendix~\ref{app:additional_experiment}, confirming that RL is an effective complement to the \alg workflow.
In this section, we also conduct experiments to investigate the following research questions: (1) Why does \alg's parallel paradigm outperform memory-based approaches? (2) How efficient is \alg in terms of inference latency? (3) How do subagent model size and document chunking affect \alg? (4) Can \alg accommodate alternative subagent implementations? All experiments in this section are conducted with Qwen3.5-4B, unless otherwise stated.

\subsection{Why {P\scalebox{0.75}{AR}S\scalebox{0.75}{ER}\xspace} Outperforms Sequential Memory Agents}
\label{sec:robustness}
To understand why \alg outperforms sequential memory agents, we construct three controlled evaluations that vary complementary aspects of evidence distribution: absolute position, logical order, and relative distance. Rather than comparing performance across methods, we focus on each method's sensitivity to perturbations applied to these three dimensions.
\begin{figure*}[ht]
\centering
\includegraphics[width=\textwidth]{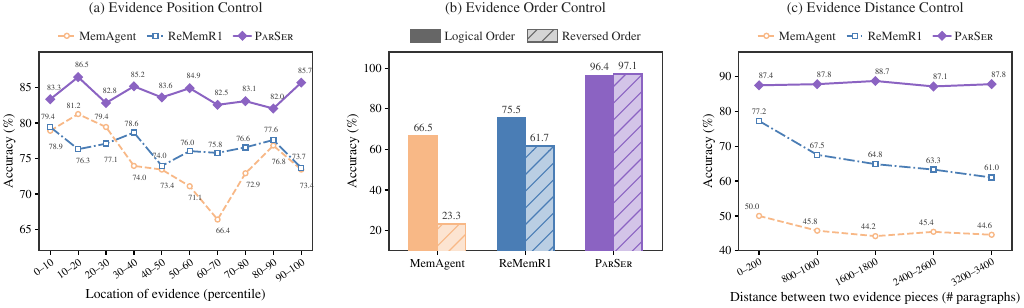}
\caption{Performance under controlled evidence distributions. (a) Evidence position: supporting evidence is placed within different percentile ranges of 894K-token documents. (b) Evidence order: evidence-bearing paragraphs in 894K-token documents either follow or reverse their logical dependency order. (c) Evidence distance: the number of intervening paragraphs between two pieces of supporting evidence is varied.}
\label{fig:evidence-controls}
\end{figure*}
\paragraph{Evidence Position Control} We manipulate the absolute position of supporting evidence within long documents. Specifically, for each of 512 test questions sampled from HotpotQA, we place all evidence-bearing paragraphs at randomly sampled positions within the $[st, st+10]$ percentile range of an 894K-token document, where $st$ ranges from 0 to 90 in increments of 10. The distractor paragraphs and their positions remain identical across variants.
Figure~\hyperref[fig:evidence-controls]{\ref*{fig:evidence-controls}(a)} shows performance across the position-controlled test sets. MemAgent suffers a pronounced performance drop when the supporting evidence lies between the 50th and 70th percentiles of the document. This is because MemAgent sequentially compresses document chunks into a fixed-capacity memory, making evidence availability dependent on where the evidence enters the memory-update sequence. Specifically, when all pieces of evidence appear near the beginning, MemAgent can aggregate them and derive an answer early; evidence near the end undergoes few subsequent memory updates. Evidence in the middle faces both prior memory saturation and subsequent overwriting, making it less likely to be retained. ReMemR1 partially mitigates this positional sensitivity by retrieving information from earlier memory states through its callback module. In contrast, \alg gives the queries symmetric access to every chunk through parallel reading, making evidence retrieval independent of position and achieving stable performance.

\paragraph{Evidence Order Control} We sample 512 two-hop bridge-comparison questions from 2WikiMultiHopQA.\footnote{We use 2WikiMultiHopQA because, unlike HotpotQA, it provides ground-truth annotations of the logical order among pieces of supporting evidence.} In this question type, the supporting evidence forms a reasoning chain in which later hops depend on entities identified in earlier hops. For example, answering the question in Figure~\ref{fig:memagent_qd_sample} requires first identifying the director of each film and then comparing the directors' dates of death. For each question, we build two 894K-token documents containing the same set of paragraphs, with all distractor paragraphs kept in the same positions. The two documents differ only in the evidence-bearing paragraphs' physical order in the document: one follows the logical dependency order, whereas the other reverses it (as in Figure~\ref{fig:memagent_qd_sample}).
Figure~\hyperref[fig:evidence-controls]{\ref*{fig:evidence-controls}(b)} compares performance under the two orders. Both MemAgent and ReMemR1 suffer a sharp performance drop when the order is reversed, whereas \alg remains stable. In sequential memory agents, when an evidence paragraph precedes the prerequisite evidence needed to recognize its relevance to the question, it may either be ignored or, if stored initially, evicted from memory before the arrival of its prerequisite (see a MemAgent example in Appendix~\ref{app:memagent_reverse}). \alg revisits all chunks under queries based on previously gathered findings, allowing the lead agent to follow the question's logical dependencies regardless of the evidence's physical order.

\paragraph{Evidence Distance Control} We study how the distance between evidence pieces affects performance. We sample 512 questions requiring two pieces of supporting evidence from 2WikiMultiHopQA. For each question, we vary their separation by inserting different numbers of distractor paragraphs between the two evidence-bearing paragraphs. To avoid the confounds identified in the two preceding experiments, we place the two evidence paragraphs following their logical order and pad a fixed set of $1{,}600$ paragraphs both before the first evidence paragraph and after the final evidence paragraph.
Figure~\hyperref[fig:evidence-controls]{\ref*{fig:evidence-controls}(c)} shows that the performance of MemAgent and ReMemR1 degrades as the number of middle paragraphs increases, whereas \alg remains stable. For sequential memory agents, the first evidence piece must survive an increasing number of memory updates before the second is encountered, making it more likely to be evicted from the fixed-capacity memory.
In \alg, the two evidence-bearing chunks are directly examined through parallel reading, and their findings are composed by the lead agent through a reasoning path independent of their physical distance.

Taken together, these experiments expose three manifestations of the same structural bottleneck in sequential memory agents: capacity-limited, document-ordered recurrent compression. 
In contrast, \alg decouples reading from reasoning into two complementary mechanisms: stateless, parallel reading provides symmetric access across the entire document regardless of evidence position or distance, while question-driven iterative reasoning aligns the inference path with the question's logical dependencies rather than the document's physical order. This separation accounts for \alg's robustness across all three controls.

\subsection{Inference Latency}
\label{sec:inference_latency}
\begin{table}[t]
    \centering
    \caption{Amortized wall-clock inference time per sample (seconds), computed as the total subset wall-clock time divided by the size of the subset (128), across document lengths under concurrency of 1, 16, and 32. Full-context methods fail to finish 896K-token evaluation at high concurrency due to GPU memory limits.}
    \label{tab:inference_latency}
    \scriptsize
    \resizebox{0.8\linewidth}{!}{%
    \begin{tabular}{llrrrrrrrr}
        \toprule
        \multirow{2}{*}{\textbf{Concurrency}} & \multirow{2}{*}{\textbf{Method}} & \multicolumn{8}{c}{\textbf{\# Paragraphs (Total Length)}} \\
        \cmidrule(l){3-10}
         & & \shortstack{50\\(7K)} & \shortstack{100\\(14K)} & \shortstack{200\\(28K)} & \shortstack{400\\(56K)} & \shortstack{800\\(112K)} & \shortstack{1600\\(224K)} & \shortstack{3200\\(448K)} & \shortstack{6400\\(896K)} \\
        \midrule
        \multirow{4}{*}{1}
            & Full-context (non-think) & 1.16 & 1.86 & 2.11 & 3.52 & 6.25 & 18.36 & 51.86 & 185.42 \\
            & Full-context (think) & 12.33 & 17.78 & 18.93 & 31.89 & 42.73 & 68.19 & 162.97 & 375.32 \\
            & MemAgent & 10.56 & 15.99 & 30.40 & 58.49 & 112.59 & 222.86 & 438.61 & 876.20 \\
            & \cellcolor{metabg}\alg & \cellcolor{metabg}5.33 & \cellcolor{metabg}5.84 & \cellcolor{metabg}6.83 & \cellcolor{metabg}8.71 & \cellcolor{metabg}13.30 & \cellcolor{metabg}22.43 & \cellcolor{metabg}42.19 & \cellcolor{metabg}78.22 \\
        \midrule
        \multirow{4}{*}{16}
            & Full-context (non-think) & 0.63 & 0.91 & 1.11 & 2.55 & 9.46 & 28.00 & 94.22 & \multicolumn{1}{c}{--} \\
            & Full-context (think) & 2.00 & 2.84 & 4.05 & 8.70 & 21.37 & 53.95 & 177.01 & \multicolumn{1}{c}{--} \\
            & MemAgent & 1.49 & 2.28 & 3.76 & 7.13 & 13.61 & 26.24 & 51.80 & 101.94 \\
            & \cellcolor{metabg}\alg & \cellcolor{metabg}0.87 & \cellcolor{metabg}1.09 & \cellcolor{metabg}1.83 & \cellcolor{metabg}3.30 & \cellcolor{metabg}7.38 & \cellcolor{metabg}15.24 & \cellcolor{metabg}31.31 & \cellcolor{metabg}58.86 \\
        \midrule
        \multirow{4}{*}{32}
            & Full-context (non-think) & 0.35 & 0.65 & 1.07 & 3.85 & 9.23 & 27.98 & 93.45 & \multicolumn{1}{c}{--} \\
            & Full-context (think) & 1.70 & 2.25 & 3.39 & 9.32 & 20.64 & 54.16 & 191.91 & \multicolumn{1}{c}{--} \\
            & MemAgent & 1.21 & 1.63 & 2.76 & 5.14 & 9.76 & 19.61 & 37.50 & 74.96 \\
            & \cellcolor{metabg}\alg & \cellcolor{metabg}0.64 & \cellcolor{metabg}0.99 & \cellcolor{metabg}1.89 & \cellcolor{metabg}3.66 & \cellcolor{metabg}7.36 & \cellcolor{metabg}15.40 & \cellcolor{metabg}30.77 & \cellcolor{metabg}58.76 \\
        \bottomrule
    \end{tabular}%
    }
\end{table}

Table~\ref{tab:inference_latency} reports three methods' amortized wall-clock inference time across all test subsets of HotpotQA with concurrency of 1, 16, and 32.\footnote{We exclude ReMemR1 from this analysis because it equips MemAgent with an extra retrieval module, which theoretically introduces higher inference latency than MemAgent.} Each entry is the total subset wall-clock time divided by the size of the subset (128), rather than the end-to-end latency of an individual request (which would typically increase under higher concurrency due to contention). Models are deployed with SGLang: we allocate one H100 GPU for Full-context and MemAgent, while \alg employs two GPUs, one H100 dedicated to all subagents and one RTX3090 for the lead agent. For \alg, the lead agent and subagents within a single inference instance run in an alternating fashion, as the lead agent has to await outputs from all subagents. Comparisons between our two-GPU \alg and single-GPU baselines are thus valid.

Across all concurrency settings, Full-context (non-thinking) achieves the lowest latency on short-document subsets because it processes the input in a single pass and generates only a short output. As document length increases, however, its single-pass processing becomes increasingly costly due to the quadratic complexity of attention module. Under high concurrency, limited GPU memory even prevents it from processing 896K-token inputs. In contrast, MemAgent and \alg process documents in fixed-length chunks, enabling them to handle longer documents within limited GPU memory.

\begin{wrapfigure}{r}{0.35\linewidth}
\centering
\includegraphics[width=\linewidth]{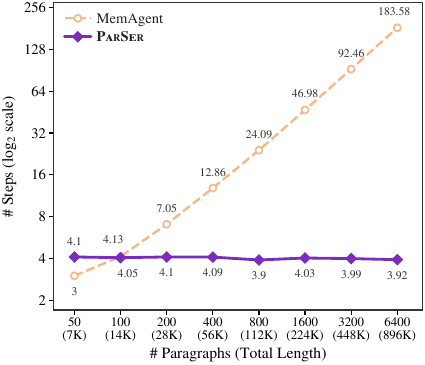}
\caption{Inference step counts of MemAgent and \alg across lengths.}
\label{fig:inference_latency_steps}
\end{wrapfigure}

Figure~\ref{fig:inference_latency_steps} compares the inference step counts of MemAgent and \alg, which help explain their latency trends. At a concurrency of 1, MemAgent's amortized latency grows linearly with document length because its number of inference steps is proportional to the number of document chunks. In contrast, \alg processes document chunks in parallel, reducing the number of sequential inference steps to the number of reasoning hops required to solve the question itself. Because the same set of questions is used across all subsets, the number of reasoning hops—and hence the inference step count of \alg—remains nearly constant even as document length increases significantly. Consequently, \alg achieves lower amortized latency than MemAgent, with an order-of-magnitude advantage on long-document subsets.
Under multi-concurrency requests, a more practical condition, MemAgent benefits from batched processing and narrows the latency gap. Nevertheless, \alg consistently maintains lower amortized time across the evaluated concurrency levels. Appendix~\ref{app:compute_analysis} explains why this advantage becomes less pronounced at higher concurrency.

\subsection{Effect of Subagent}
\label{sec:subagent_effect}
\begin{table}[t]
    \centering
    \caption{Effect of subagent size on HotpotQA (Sub\_EM, \%). The lead agent is fixed as Qwen3.5-4B. Best results in each column are \textbf{bolded}.}
    \label{tab:analysis_subagent}
    \scriptsize
    \resizebox{0.81\linewidth}{!}{%
    \begin{tabular}{lccccccccc}
        \toprule
        \multirow{2}{*}{\textbf{Subagent}} & \multicolumn{8}{c}{\textbf{\# Paragraphs (Total Length)}} & \multirow{2}{*}{\textbf{Avg.}} \\
        \cmidrule(lr){2-9}
         & \shortstack{50\\(7K)} & \shortstack{100\\(14K)} & \shortstack{200\\(28K)} & \shortstack{400\\(56K)} & \shortstack{800\\(112K)} & \shortstack{1600\\(224K)} & \shortstack{3200\\(448K)} & \shortstack{6400\\(896K)} & \\
        \midrule
        Qwen3.5-2B & 82.55 & 84.38 & 82.29 & 84.11 & 77.87 & 76.56 & 69.01 & 69.27 & 78.26 \\
        Qwen3.5-4B (default) & \textbf{85.68} & 84.64 & 83.60 & \textbf{85.68} & \textbf{85.42} & 83.07 & \textbf{83.07} & 85.42 & 84.57 \\
        Qwen3.5-9B & 82.81 & \textbf{85.16} & \textbf{87.50} & 84.38 & 84.38 & \textbf{85.16} & 82.81 & \textbf{85.94} & \textbf{84.77} \\
        \bottomrule
    \end{tabular}%
    }
\end{table}

\paragraph{Subagent Size}
Besides Qwen3.5-4B used by default, we add 2B and 9B models as alternative subagents without retraining the lead agents. In Table~\ref{tab:analysis_subagent}, performance improves when replacing the 2B subagent with the 4B version, and then saturates when employing an even larger subagent (9B). We attribute this saturation to the simplification of the subagents' task. After question decomposition and document chunking, each subagent only needs to answer a focused query over a short context, for which the 4B model already provides sufficient capacity. This highlights the deployment efficiency of \alg via the adoption of lightweight subagents.

\begin{table}[t]
    \centering
    \caption{Effect of chunk size on HotpotQA (Sub\_EM, \%). All variants are based on Qwen3.5-4B. For the Full-document setting, YaRN-based RoPE scaling is enabled on the 3200~(448K) and 6400~(896K) subsets to accommodate contexts beyond the model's native window.}
    \label{tab:analysis_chunk_size}
    \scriptsize
    \resizebox{0.75\linewidth}{!}{%
    \begin{tabular}{lccccccccc}
        \toprule
        \multirow{2}{*}{\textbf{Chunk size}} & \multicolumn{8}{c}{\textbf{\# Paragraphs (Total Length)}} & \multirow{2}{*}{\textbf{Avg.}} \\
        \cmidrule(lr){2-9}
         & \shortstack{50\\(7K)} & \shortstack{100\\(14K)} & \shortstack{200\\(28K)} & \shortstack{400\\(56K)} & \shortstack{800\\(112K)} & \shortstack{1600\\(224K)} & \shortstack{3200\\(448K)} & \shortstack{6400\\(896K)} & \\
        \midrule
        4096 & 85.68 & 84.64 & 83.60 & 85.68 & 85.42 & 83.07 & 83.07 & 85.42 & 84.57 \\
        Full & 83.85 & 84.64 & 82.29 & 83.86 & 74.48 & 71.10 & 56.77 & 53.13 & 73.76 \\
        16{,}384 & 82.03 & 82.29 & 85.94 & 82.81 & 82.81 & 83.86 & 85.42 & 83.59 & 83.59 \\
        65{,}536 & 84.64 & 82.03 & 80.73 & 80.73 & 78.91 & 80.99 & 83.59 & 83.08 & 81.84 \\
        131{,}072 & 83.07 & 83.59 & 80.99 & 80.21 & 75.78 & 80.21 & 79.69 & 79.17 & 80.34 \\
        \bottomrule
    \end{tabular}%
    }
\end{table}

\paragraph{Chunk Size}
The subagents of \alg use the same working mode as the Full-context (non-thinking) baseline introduced in \S\ref{sec:baselines}, but receive inputs at a different document granularity (document chunk vs. full  document). Thus, the improvement of \alg over Full-context (non-thinking) in Table~\ref{tab:main_hqa} can be attributed to both the lead agent's guidance and the chunk-level document decomposition. To isolate their respective contributions, we introduce an ablated \alg variant without chunking, which employs a single subagent fed with the full document. Table~\ref{tab:analysis_chunk_size} reports a notable performance drop upon the removal of chunking, particularly on long-document subsets. We also add variants with intermediate chunk sizes and observe that performance drops as chunk size increases. This suggests that key information contained in longer input is more difficult for LLMs to capture than in shorter input, consistent with the \textit{context rot} phenomenon observed in prior work \citep{hong2025context, liu-etal-2024-lost}.
To mitigate this issue, \alg instantiates multiple subagents, and each subagent processes only a short chunk of the document.

\subsection{Compatibility with Alternative Subagent Implementations}
\label{sec:subagent_compatibility}
\begin{table}[t]
    \centering
    \caption{Comparison of alternative subagent implementations with their full-context counterparts on HotpotQA (Sub\_EM, \%). All agents are built on Qwen3.5-4B.}
    \label{tab:analysis_generalizability_subagents}
    \scriptsize
    \resizebox{0.84\linewidth}{!}{%
    \begin{tabular}{lccccccccc}
        \toprule
        \multirow{2}{*}{\textbf{Agent}} & \multicolumn{8}{c}{\textbf{\# Paragraphs (Total Length)}} & \multirow{2}{*}{\textbf{Avg.}} \\
        \cmidrule(lr){2-9}
         & \shortstack{50\\(7K)} & \shortstack{100\\(14K)} & \shortstack{200\\(28K)} & \shortstack{400\\(56K)} & \shortstack{800\\(112K)} & \shortstack{1600\\(224K)} & \shortstack{3200\\(448K)} & \shortstack{6400\\(896K)} & \\
        \midrule
        DCI agent & 73.96 & 75.78 & 74.74 & 76.56 & 74.48 & 75.52 & 77.60 & 76.30 & 75.61 \\
        Lead agent + DCI subagents & \textbf{86.72} & \textbf{88.28} & \textbf{89.06} & \textbf{86.72} & \textbf{85.94} & \textbf{80.47} & \textbf{81.25} & \textbf{78.91} & \textbf{84.67} \\
        \midrule
        Full-context (think) & 80.47 & 78.91 & 78.12 & 76.56 & 74.22 & 60.94 & 46.09 & 31.25 & 65.82 \\
        Lead agent + thinking subagents & \textbf{85.16} & \textbf{86.72} & \textbf{87.50} & \textbf{82.03} & \textbf{85.94} & \textbf{85.94} & \textbf{87.50} & \textbf{83.59} & \textbf{85.55} \\
        \bottomrule
    \end{tabular}%
    }
\end{table}

\alg defaults to direct-answer subagents. We investigate whether the lead agent is compatible with alternative subagent implementations without retraining. We add two additional subagent variants: the thinking subagents and the DCI subagents. DCI (Direct Corpus Interaction; \citealt{li2026beyond}, \citealt{salemi2026grepseektrainingsearchagents}, \citealt{sen2026grepneedagentharnesses}) refers to an agentic search paradigm that enables LLMs to directly query raw text corpora via composable Unix shell tools like \texttt{rg} and \texttt{grep} for fine-grained lexical matching and multi-step evidence collection. We follow the implementation of \cite{li2026beyond} to build our DCI subagents, whose implementation details are presented in Appendix~\ref{app:dci_details}. As shown in Table~\ref{tab:analysis_generalizability_subagents}, both variants outperform their corresponding standalone baselines: \alg with thinking subagents improves over Full-context (think), while \alg with DCI subagents improves over the standalone DCI agent. These results suggest that the lead agent can effectively coordinate different subagent implementations.

\section{Related Work}
\label{sec:related}

\subsection{Long-Context LLMs}

Supporting million-token contexts \emph{efficiently} has driven two complementary lines of architectural work. Positional interpolation rescales rotary embeddings so a model trained on short sequences extrapolates to far longer ones \citep{chen2023pi, peng2024yarn, ding2024longrope}. A separate family of attention mechanisms attacks the quadratic cost that dominates at long context: sparse attention attends only to a learned subset of relevant tokens per query \citep{deepseekai2025nsa, deepseekv32, minimax2026sparse}, and linear-attention variants reduce the complexity to linear in sequence length \citep{kimiteam2026k3}. Yet a longer window does not by itself yield better \emph{use} of that window. Models systematically underuse evidence placed in the middle of their input \citep{liu-etal-2024-lost}, and accuracy degrades as the input grows even when the nominal window is far from full, a phenomenon documented as \emph{context rot} \citep{hong2025context}. A prominent response sidesteps the window limit altogether by reading the document in chunks while maintaining a compact textual memory that is repeatedly rewritten. Training-free methods precompute and then navigate such a memory. ReadAgent \citep{lee2024readagent} uses gist lookup and MemWalker \citep{chen2023memwalker} uses a summary tree, while Chain-of-Agents \citep{zhang2024chainofagents} assigns one chunk per worker but passes a single message \emph{sequentially} down the chain. MemAgent \citep{yu2026memagent} instead trains this recurrent read-and-compress workflow end-to-end with reinforcement learning. ReMemR1 \citep{shi2026look} adds a callback that revisits earlier memory states, and GRU-Mem \citep{sheng2026grumem} introduces gated updates with an early-exit mechanism. These methods share one commitment: processing $T$ chunks requires $T$ \emph{dependent} steps. Three consequences follow. Latency grows linearly with document length, a fixed-capacity memory must irreversibly decide what to retain before downstream relevance can be known, and the outcome depends on the order in which evidence is encountered \citep{gupta2026chowliu}. \secref{sec:robustness} confirms that these are measurable biases with respect to evidence position, order, and separation.

\subsection{Parallel Reading via Orchestrator–Worker Architectures}

Reading chunks independently and aggregating their results is the natural parallel alternative to a sequential memory. Map-reduce pipelines such as LLM$\times$MapReduce \citep{zhou2024llmxmapreduce} and ToM \citep{guo2025tom} explore this direction, but most are \emph{single-shot}: the query sent to each chunk is fixed before any chunk is read. This cannot handle multi-hop questions, where later hops are not recognizable until earlier ones are found \citep{xu2026divideconquer}. LongAgent \citep{zhao2024longagent} and XpandA \citep{xpanda2025} pair a leader with per-chunk agents over multiple rounds, but coordinate through hand-specified protocols rather than a learned policy. A separate group achieves parallelism \emph{inside} the model by encoding chunks independently and fusing them at the attention level \citep{ratner2023pcw, merth2024superposition, ma2025blockattention, yang2025ape, yen2024cepe}, but these are query-agnostic, single-round, and require architecture surgery. Structurally, a leader dispatching subtasks to workers that each run in an isolated context window is by now a common pattern in agentic systems \citep{anthropic2025multiagent}, since a worker's intermediate tokens never occupy the leader's context. A growing line of work trains \emph{only} this orchestrator while keeping the workers frozen \citep{hu2025owl, dang2025puppeteer}, a design echoed by commercial agent swarms that optimize the scheduler alone \citep{kimi2025k25}. \alg is the \emph{long-context instantiation} of this design. Its subagents are bound to a disjoint partition of the input, so coverage is guaranteed by construction and the lead agent's task reduces to query formulation and aggregation. This structure makes freezing the subagents viable (\secref{sec:subagent_compatibility}) and keeps training cost independent of document length. 

\section{Conclusion}
\label{sec:conclusion}

We presented \alg, a long-context reasoning framework that decouples reading from reasoning: frozen subagents read chunks in parallel, while an RL-trained lead agent iteratively refines queries and aggregates evidence. Across contexts from $7$K to $896$K tokens, \alg outperforms the strongest sequential memory baselines by $5.7$--$6.7$ points on average (up to $12.0$ at $896$K), while its 9B variant surpasses the million-token full-context DeepSeek-V4-Pro by $6.3$ points. It also reduces latency while remaining robust to context length and evidence placement.

\clearpage
\newpage
\bibliographystyle{assets/plainnat}
\bibliography{paper}

\clearpage
\newpage
\beginappendix
\section{Time Complexity Analysis}
\label{app:compute_analysis}
We analyze  the time complexity of Full-context (direct answering given the entire document), MemAgent, and \alg. ReMemR1 operates in a pipeline highly similar to MemAgent, so they have the same level of time complexity. Given a question and a document with $n$ tokens, Full-context and MemAgent generate responses with $r_F$ and $r_M$ tokens, respectively. For \alg, the lead agent and all subagents generate $r_L$ and $r_S$ tokens in total, respectively. For MemAgent and \alg, each document is split into $T$ chunks.
As documents have significantly more tokens than the concatenation of task instructions and questions, the effective input sequence length for all three approaches can be approximated as $n$. To simplify the complexity derivation, we assume \textit{the GPU memory capacity is sufficient to hold the entire contextual KV cache}, thereby eliminating recomputation of context's key-value representations.

\begin{table}[t]
    \setlength{\belowcaptionskip}{6pt}
    \centering
    \caption{Average output token counts per sample under different document lengths. Both methods are implemented based on Qwen3.5-4B. MemAgent entries report the total output tokens. For \alg, entries report the combined output-token counts of the lead agent and all subagents. The final row reports the ratio of \alg to MemAgent theoretical decoding computation, calculated from Equations \eqref{eq:memagent_decode} and \eqref{eq:longmas_decode} with $n/T=5000$ and $K=4$, under the assumption of sufficient GPU memory to retain all chunk KV caches.}
    \label{tab:output_tokens}
    \scriptsize
    \resizebox{\linewidth}{!}{%
    \begin{tabular}{lrrrrrrrr}
        \toprule
        & \multicolumn{8}{c}{\textbf{\# Paragraphs (Total Length)}} \\
        \cmidrule(l){2-9}
        & \shortstack{50\\(7K)} & \shortstack{100\\(14K)} & \shortstack{200\\(28K)} & \shortstack{400\\(56K)} & \shortstack{800\\(112K)} & \shortstack{1600\\(224K)} & \shortstack{3200\\(448K)} & \shortstack{6400\\(896K)} \\
        \midrule
        \shortstack[l]{MemAgent\\\quad(\# tokens)} & 2211 & 3379 & 6353 & 12334 & 23805 & 47174 & 93806 & 187058 \\
        \shortstack[l]{\alg\\\quad(\# tokens by Lead + Subagents)} & 534+312 & 533+348 & 525+379 & 485+392 & 523+474 & 522+515 & 546+586 & 541+804 \\
        \shortstack[l]{\alg/MemAgent\\\quad(Decode Computation)} & 13.9\% & 10.6\% & 6.1\% & 3.2\% & 2.0\% & 1.1\% & 0.6\% & 0.4\% \\
        \bottomrule
    \end{tabular}%
    }
    \vspace{-8pt}
\end{table}

\paragraph{Prefill} Owing to the inherent properties of the attention mechanism in Transformer architectures \citep{3295222.3295349}, the Full-context baseline exhibits the highest prefill-phase time complexity $\mathcal{O}(n^2)$, which scales quadratically with the input sequence length. Through the adoption of chunk-level input, MemAgent and \alg reduce the prefill-phase time complexity as
\begin{equation}
    T \cdot \mathcal{O}\left(\left(\frac{n}{T}\right)^2\right)
    = \mathcal{O}\left(\frac{n^2}{T}\right),
    \label{eq:prefill_complexity}
\end{equation}
where each of the $T$ chunks contains $n/T$ tokens and is encoded independently.

\paragraph{Decode} Under cached autoregressive decoding, each newly generated token attends to all preceding input and output tokens. Thus, decoding $r_F$ tokens from the $n$-token Full-context input has time complexity
\begin{equation}
    \mathcal{O}\left(\sum_{t=1}^{r_F}(n+t)\right)
    = \mathcal{O}\left(nr_F+r_F^2\right).
\end{equation}
Based on the operation pipeline of MemAgent, for each chunk, the model generates an average of $r_M/T$ tokens for memory update. Its decoding complexity is therefore
\begin{equation}
    T \cdot \mathcal{O}\left(\frac{n}{T}\frac{r_M}{T}+\left(\frac{r_M}{T}\right)^2\right)
    = \mathcal{O}\left(\frac{nr_M+r_M^2}{T}\right).
    \label{eq:memagent_decode}
\end{equation}

For \alg, let $K$ denote the number of lead-agent rounds and $r'_S$ the average number of tokens generated by each of the $T$ subagents per round. Thus, the total number of tokens generated by all subagents is $r_S=KTr'_S$. In each round, the $T$ subagents independently decode over their respective $n/T$-token chunks. Their aggregate decoding computation is
\begin{equation}
    KT \cdot \mathcal{O}\left(\frac{n}{T}r'_S+\left(r'_S\right)^2\right)
    = \mathcal{O}\left(Knr'_S+KT\left(r'_S\right)^2\right).
\end{equation}
The lead agent receives an observation of $r_S$ tokens generated by subagents and generates $r_L$ tokens. Its decoding complexity is $\mathcal{O}(r_Sr_L+r_L^2)$. Consequently, the total decoding computation of \alg is
\begin{equation}
    \mathcal{O}\left(Knr'_S+KT\left(r'_S\right)^2+r_Sr_L+r_L^2\right).
\end{equation}
Using the total subagent output $r_S=KTr'_S$, this complexity can equivalently be written as
\begin{equation}
    \mathcal{O}\left(\frac{nr_S}{T}+\frac{r_S^2}{KT}+r_Sr_L+r_L^2\right).
    \label{eq:longmas_decode}
\end{equation}
When the $T$ subagents are executed in parallel, their decoding contribution to wall-clock latency is reduced by a factor of $T$. The corresponding latency is
\begin{equation}
    \mathcal{O}\left(\frac{nr_S}{T^2}+\frac{r_S^2}{KT^2}+r_Sr_L+r_L^2\right).
\end{equation}
Based on the decoding time-complexity formulas in \eqref{eq:memagent_decode} and \eqref{eq:longmas_decode}, we compute the ratio of \alg to MemAgent decoding computation under each document length; the resulting ratios are reported in the last row of Table~\ref{tab:output_tokens}. The large gap arises mainly from the disparity in total generated tokens: MemAgent produces a memory of about 1K tokens for every chunk, whereas each \alg subagent emits only a short finding or an ``Unknown'' response for its chunk---and the latter dominates in most cases. As a result, MemAgent's output length grows roughly with the number of chunks, while \alg's remains comparatively small and stable, yielding substantially lower decoding complexity.
Moreover, \alg exposes parallelism across the $T$ independent subagents. With sufficient hardware resources, this reduces the subagent contribution to wall-clock decoding latency from $\mathcal{O}(nr_S/T+r_S^2/(KT))$ to $\mathcal{O}(nr_S/T^2+r_S^2/(KT^2))$, while the lead-agent terms remain sequential.

Note that this analysis assumes that GPU memory is sufficient to retain the KV caches for all encoded chunks. When this assumption does not hold, especially at high request concurrency, subagents' chunk KV caches may be evicted. The subagents must then re-prefill their chunks in each query round, increasing the aggregate prefill complexity from $\mathcal{O}(n^2/T)$ to $\mathcal{O}(Kn^2/T)$. With $K=4$, as measured in \S\ref{sec:inference_latency}, this is four times the prefill computation of MemAgent. In contrast, MemAgent updates memory by sequentially scanning chunks and does not revisit previously processed chunks; thus, it is unaffected by this KV-cache retention constraint. This explains why \alg's latency advantage is less pronounced at higher concurrency, as shown in \S\ref{sec:inference_latency}. 

\section{Additional Experiments}
\label{app:additional_experiment}
\subsection{Efficacy of Reinforcement Learning}

Table~\ref{tab:rl_hqa} compares MemAgent, ReMemR1, and \alg with and without RL training. Within the \alg group, RL brings a clear gain: average HotpotQA accuracy rises from $74.42$ to $84.57$ on Qwen3.5-4B and from $76.24$ to $86.79$ on Qwen3.5-9B, with a smaller but consistent lift on out-of-distribution 2WikiMultiHopQA ($82.85\!\rightarrow\!87.04$ and $84.86\!\rightarrow\!88.48$). These gaps confirm that RL is an effective complement to \alg.

Even without RL, \alg already outperforms MemAgent and ReMemR1 by a wide margin and remains essentially flat as document length grows, whereas the sequential memory agents degrade sharply. We attribute this to a closer match between \alg and the pretrained model: the lead agent follows the model's native multi-turn tool-calling template, a format the backbone is already trained to follow, whereas sequential memory agents impose a custom recurrent memory-update interface that the pretrained checkpoint has never seen.

\newcommand{\lightmid}{%
    \arrayrulecolor{black!25}%
    \cmidrule(lr){2-12}%
    \arrayrulecolor{black}%
}

\begin{table}[t]
    \centering
    \caption{Effect of reinforcement learning on HotpotQA~\citep{yang2018hotpotqa} and 2WikiMultiHopQA~\citep{xanh2020_2wikimultihop}. Values are accuracy (Sub\_EM, \%).}
    \label{tab:rl_hqa}
    \label{tab:rl_2wiki}
    \scriptsize
    \resizebox{0.9\linewidth}{!}{%
    \begin{tabular}{ll@{\ }lccccccccc}
        \multicolumn{12}{c}{(a) Accuracy on HotpotQA (In-Distribution)} \\
        \midrule
        \multirow{2}{*}{\textbf{Backbone}} & \multirow{2}{*}{\textbf{Method}} & & \multicolumn{8}{c}{\# \textbf{Paragraphs (Total Length)}} & \multirow{2}{*}{\textbf{Avg.}} \\
        \cmidrule(lr){4-11}
         & \multicolumn{2}{c}{} & \shortstack{50\\(7K)} & \shortstack{100\\(14K)} & \shortstack{200\\(28K)} & \shortstack{400\\(56K)} & \shortstack{800\\(112K)} & \shortstack{1600\\(224K)} & \shortstack{3200\\(448K)} & \shortstack{6400\\(896K)} & \\
        \midrule
        \multirow{6}{*}{Qwen3.5-4B}
            & MemAgent & w/o RL & 69.53 & 62.76 & 58.86 & 50.78 & 40.62 & 32.29 & 25.78 & 19.79 & 45.05 \\
            &          & w/ RL  & 81.25 & 81.51 & 83.59 & 77.86 & 77.86 & 79.69 & 73.96 & 72.92 & 78.58 \\
        \lightmid
            & ReMemR1  & w/o RL & 66.41 & 63.28 & 59.90 & 48.44 & 44.53 & 34.38 & 27.87 & 20.58 & 45.67 \\
            &          & w/ RL  & 82.03 & 79.95 & 82.03 & 78.91 & 77.86 & 79.95 & 77.08 & 73.44 & 78.91 \\
        \lightmid
            & \alg     & w/o RL & 72.92 & 76.04 & 73.96 & 74.48 & 76.04 & 73.96 & 75.00 & 72.92 & 74.42 \\
            &          & w/ RL  & 85.68 & 84.64 & 83.60 & 85.68 & 85.42 & 83.07 & 83.07 & 85.42 & 84.57 \\
        \midrule
        \multirow{6}{*}{Qwen3.5-9B}
            & MemAgent & w/o RL & 68.49 & 61.20 & 53.64 & 52.08 & 47.14 & 38.80 & 26.82 & 26.30 & 46.81 \\
            &          & w/ RL  & 81.77 & 80.73 & 82.03 & 79.95 & 79.95 & 81.25 & 79.69 & 75.00 & 80.05 \\
        \lightmid
            & ReMemR1  & w/o RL & 61.20 & 53.64 & 46.10 & 46.35 & 42.71 & 39.58 & 40.62 & 32.03 & 45.28 \\
            &          & w/ RL  & 81.25 & 79.69 & 77.60 & 78.39 & 78.13 & 78.13 & 77.86 & 76.04 & 78.39 \\
        \lightmid
            & \alg     & w/o RL & 77.60 & 74.22 & 74.48 & 74.74 & 79.43 & 77.34 & 76.56 & 75.52 & 76.24 \\
            &          & w/ RL  & 86.72 & 88.80 & 87.24 & 85.68 & 86.46 & 86.72 & 86.72 & 85.94 & 86.79 \\
        \midrule
        \multicolumn{12}{c}{\shortstack{\strut\\(b) Accuracy on 2WikiMultiHopQA (Out-of-Distribution)}} \\
        \midrule
        \multirow{2}{*}{\textbf{Backbone}} & \multirow{2}{*}{\textbf{Method}} & & \multicolumn{8}{c}{\# \textbf{Paragraphs (Total Length)}} & \multirow{2}{*}{\textbf{Avg.}} \\
        \cmidrule(lr){4-11}
         & \multicolumn{2}{c}{} & \shortstack{50\\(7K)} & \shortstack{100\\(14K)} & \shortstack{200\\(28K)} & \shortstack{400\\(56K)} & \shortstack{800\\(112K)} & \shortstack{1600\\(224K)} & \shortstack{3200\\(448K)} & \shortstack{6400\\(896K)} & \\
        \midrule
        \multirow{6}{*}{Qwen3.5-4B}
            & MemAgent & w/o RL & 72.92 & 72.66 & 62.76 & 59.63 & 46.61 & 44.01 & 42.71 & 34.90 & 54.53 \\
            &          & w/ RL  & 67.45 & 73.18 & 69.01 & 61.46 & 54.17 & 54.43 & 60.16 & 45.05 & 60.61 \\
        \lightmid
            & ReMemR1  & w/o RL & 77.86 & 71.88 & 60.68 & 56.25 & 52.08 & 46.35 & 45.31 & 36.98 & 55.93 \\
            &          & w/ RL  & 88.54 & 86.85 & 83.20 & 83.07 & 76.83 & 67.71 & 72.40 & 60.68 & 77.41 \\
        \lightmid
            & \alg     & w/o RL & 82.29 & 84.38 & 85.68 & 84.90 & 80.99 & 81.51 & 83.33 & 79.69 & 82.85 \\
            &          & w/ RL  & 86.98 & 84.64 & 88.28 & 87.76 & 87.50 & 85.94 & 88.28 & 86.98 & 87.04 \\
        \midrule
        \multirow{6}{*}{Qwen3.5-9B}
            & MemAgent & w/o RL & 74.48 & 69.79 & 60.94 & 60.42 & 50.26 & 44.53 & 44.79 & 31.51 & 54.59 \\
            &          & w/ RL  & 80.73 & 82.03 & 78.91 & 75.26 & 70.57 & 65.63 & 75.78 & 60.94 & 73.73 \\
        \lightmid
            & ReMemR1  & w/o RL & 77.34 & 71.09 & 61.20 & 63.28 & 55.21 & 50.52 & 48.96 & 40.89 & 58.56 \\
            &          & w/ RL  & 84.12 & 87.50 & 77.87 & 83.59 & 76.30 & 74.74 & 79.43 & 70.57 & 79.27 \\
        \lightmid
            & \alg     & w/o RL & 84.37 & 86.46 & 86.98 & 83.07 & 83.59 & 83.86 & 85.94 & 84.64 & 84.86 \\
            &          & w/ RL  & 87.76 & 88.80 & 88.80 & 89.06 & 87.24 & 89.58 & 88.54 & 88.02 & 88.48 \\
        \bottomrule
    \end{tabular}%
    }
\end{table}

\subsection{Out-of-Distribution Evaluation on RULER}
\label{app:ruler_ood}

To examine whether \alg generalizes beyond the multi-hop QA formats used for training, we evaluate it on ten synthetic tasks from the RULER benchmark~\citep{hsieh2024ruler}. We directly use the test instances constructed by MemAgent~\citep{yu2026memagent}, which cover context lengths from 8K to 512K tokens. The tasks span three long-context capabilities. The needle-in-a-haystack (NIAH) suite evaluates retrieval: the three single-key variants retrieve one target key--value pair under different distractor densities; the three multi-key variants retrieve a target among multiple distractor needles; multi-value requires extracting all values associated with the same key; and multi-query requires answering multiple distinct keys. Variable Tracking evaluates multi-hop tracing over chains of entity assignments, while Frequent Words Extraction evaluates aggregation by asking for the most frequent words in a power-law word distribution.

We evaluate both the 4B and 9B \alg lead agents. For both variants, every chunk-bound subagent uses Qwen3.5-4B with a chunk size of $8{,}192$ tokens. All other inference settings follow the main experiments.

\begin{figure*}[t]
    \centering
    \includegraphics[width=\textwidth]{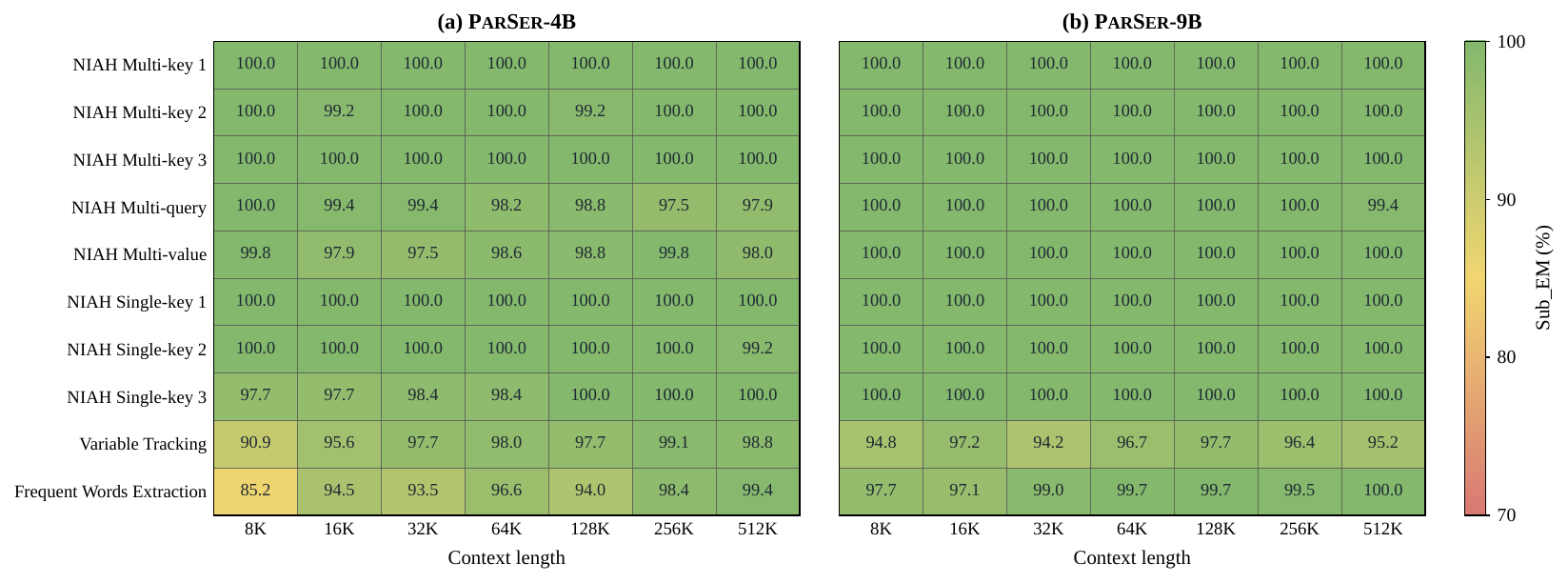}
    \caption{Out-of-distribution performance of the 4B and 9B \alg on ten RULER tasks across context lengths from 8K to 512K tokens. Each cell reports Sub\_EM (\%).}
    \label{fig:ruler_ood}
\end{figure*}

As shown in Figure~\ref{fig:ruler_ood}, \alg achieves consistently high Sub\_EM across context lengths on most tasks. Interestingly, for VT and FWE, the performance of the 4B lead agent improves rather than degrades as the context grows. Inspection of these cases suggests that longer context distributes the fixed number of key information pieces across more individual chunks, and the independently operating subagents can then surface more pieces from their respective chunks. 
\section{Datasets}
\label{app:data_details}

\subsection{Training Data Construction}
We construct the training set by following Stage~I of MemAgent \citep{yu2026memagent}; ReMemR1 \citep{shi2026look} adopts the same Stage~I recipe. We do \emph{not} use MemAgent's Stage~II data.

Concretely, we start from HotpotQA \citep{yang2018hotpotqa} training questions and retain each question's supporting Wikipedia articles as gold evidence. Following the RULER \citep{hsieh2024ruler}-style packing used by MemAgent \citep{yu2026memagent}, we pad each sample with distractor articles sampled from the same HotpotQA corpus until the context contains $200$ paragraphs ($\sim28$K tokens), then shuffle the paragraph order with a fixed random seed. To remove questions that are already solvable from parametric knowledge alone, we query Qwen3.5-9B \citep{qwen35blog} in the non-thinking mode \emph{without} providing any document context, sample Best-of-3 responses, and discard any question for which the model achieves a $100\%$ score under the rule-based verifier (substring / boxed-answer matching as in MemAgent). $41{,}027$ HotpotQA training examples are processed through this pipeline; we take the first $32{,}768$ remaining samples as our RL training set.

\subsection{Evaluation Data Construction}
\paragraph{HotpotQA (in-distribution).}
We directly reuse the long-context HotpotQA evaluation sets released by MemAgent \citep{yu2026memagent}\footnote{\url{https://github.com/BytedTsinghua-SIA/MemAgent}}. MemAgent synthesizes $128$ questions from the HotpotQA validation split with the same packing recipe as training, then provides each question with contexts of $L\in\{50,100,200,400,800,1600,\\3200,6400\}$ paragraphs (approximately $7$K--$896$K tokens), reusing the same question indices across all length settings.

\paragraph{2WikiMultiHopQA (out-of-distribution).}
For out-of-distribution evaluation we use 2WikiMultiHopQA \citep{xanh2020_2wikimultihop}. Because ReMemR1 \citep{shi2026look} does not release the constructed test files, we regenerate them with the authors' public data-processing script\footnote{\url{https://github.com/syr-cn/ReMemR1}}. The script loads 2WikiMultiHopQA from FlashRAG \citep{FlashRAG}, retains supporting-fact evidence for each question, pads the context with random distractor paragraphs to the same $L$ grid as above, shuffles paragraphs and keeps $128$ samples per length setting.

Table~\ref{tab:data_stats} summarizes the training and evaluation data. For each evaluation benchmark, the notation $128\times 8$ indicates $128$ shared questions evaluated under the eight paragraph-count settings listed in the last column.

\begin{table}[t]
    \centering
    \caption{Statistics of training and evaluation data. Training contexts use $200$ paragraphs ($\sim28$K tokens). Evaluation contexts range from $50$ to $6400$ paragraphs ($\sim7$K--$896$K tokens), with the same $128$ questions reused across the eight length settings.}
    \label{tab:data_stats}
    \scalebox{0.8}{%
    \begin{tabular}{lcc}
        \toprule
        \textbf{Split} & \textbf{\# Samples} & \textbf{\# Paragraphs per sample} \\
        \midrule
        Training (HotpotQA) & $32{,}768$ & $200$ \\
        Eval (HotpotQA) & $128\times 8$ & $\{50,100,200,400,800,1600,3200,6400\}$ \\
        Eval (2WikiMultiHopQA) & $128\times 8$ & $\{50,100,200,400,800,1600,3200,6400\}$ \\
        \bottomrule
    \end{tabular}%
    }
\end{table}

\section{Baseline Implementation Details}
\label{app:baseline_implementation}
\subsection{Full-Context Answering}
\label{app:full_context_details}
Full-context answering feeds the question together with the entire associated document into a single LLM call and asks the model to answer directly. We evaluate Qwen3.5 \citep{qwen35blog} under this protocol (both thinking and non-thinking modes). For inputs within the model's native $262$K-token window we use the default configuration; for longer documents we apply YaRN-based RoPE scaling \citep{peng2024yarn} with a factor of $4.0$, extending the effective context to about $1$M tokens. Generation uses temperature $0$. DeepSeek-V4-Pro (preview, 2026-04-24) \citep{deepseekai2026deepseekv4} natively supports a $1$M-token context window; we evaluate it under the same full-context input format with its reasoning effort set to \texttt{Max}.

Each sample is answered in two turns: the user prompt in the first turn provides the full document and the question, and the models perform inference optionally with thinking; the second turn is a short follow-up that asks for a concise final answer only. We adopt this protocol because thinking models fail to strictly follow format instructions after finishing their reasoning more frequently than non-thinking models (e.g., wrapping the final answer in designated tags) \citep{li2025when}; the issue becomes more pronounced after lengthy deliberation over long documents. In a preliminary experiment that required placing the final answer inside \texttt{<answer>} and \texttt{</answer>}, DeepSeek-V4-Pro under \texttt{think-max} violated the format instruction on $9.47\%$ of in-distribution evaluation cases. We therefore separate reasoning from answer presentation in a two-turn format.

\begin{promptbox}{System prompt:}
\small\ttfamily
You are a helpful assistant that answers complex multi-hop questions based on the provided reference documents.

\par\vspace{0.5\baselineskip}
Read the documents carefully and give an answer based on the documents.
\end{promptbox}

\begin{promptbox}{User prompt for the first turn:}
\small\ttfamily
\#\# Reference Document

\par\vspace{0.5\baselineskip}
\texttt{\{context\}}

\par\vspace{0.5\baselineskip}
\#\# Question

\par\vspace{0.5\baselineskip}
\texttt{\{question\}}
\end{promptbox}

\begin{promptbox}{User prompt for the second turn:}
\small\ttfamily
What is your final answer? Give only a concise answer. Do not include reasoning or any other text.
\end{promptbox}

\subsection{Direct Corpus Interaction}
\label{app:dci_details}
We reimplement Direct Corpus Interaction (DCI; \citealt{li2026beyond,salemi2026grepseektrainingsearchagents,sen2026grepneedagentharnesses}) as a single-agent long-document QA baseline. Unlike other approaches that ingest document chunks into the context, DCI keeps the corpus outside the conversation and lets the model inspect it only through local shell tools.

We mainly follow the implementation of DCI in \citet{li2026beyond}.\footnote{\url{https://github.com/DCI-Agent/DCI-Agent-Lite}} The agent is equipped with two native tools: (i)~``\texttt{read}'', which returns a line-numbered slice of a corpus file with a default window of $200$ lines; and (ii)~``\texttt{bash}'', which executes a read-only shell command in the corpus directory (primarily \texttt{rg}, together with ordinary inspection utilities such as \texttt{ls}, \texttt{find}, \texttt{head}, and \texttt{wc}). Parallel tool calls within a single model turn are allowed. Bash execution is confined to a sandbox. Tool observations are truncated to at most $20{,}000$ characters (\texttt{read} keeps the head; \texttt{bash} keeps the tail). When the cumulative size of tool results exceeds $240{,}000$ characters, we apply the zero-LLM L3 history compaction of \citet{li2026beyond}, replacing older tool results with a short placeholder while retaining the most recent $12$ results.

Inference proceeds in a multi-turn ReAct-style loop with a maximum of $48$ turns. We enable the model's thinking mode, set temperature to $0$, and cap each generation at $8{,}192$ tokens; length-truncated turns receive a continuation observation and continue. When the model stops without tool calls, its response is taken as the final answer. If the turn budget is exhausted before a usable answer appears, we append a tool-free finalize prompt that asks the model to emit a concise answer from evidence already present in the trajectory. All DCI baselines are served with SGLang under the same Qwen3.5 backbone family as \alg.

\subsection{Agentic RAG}
\label{app:rag_details}
We implement Agentic RAG as a multi-turn retrieval baseline that uses the same Qwen3.5 backbone family \citep{qwen35blog} and tool-calling interface as \alg, but replaces the document-reading subagents with a dense retriever. The agent is given only the question at initialization; the corpus remains outside its context until it invokes \texttt{retrieve\_documents} with a natural-language query. The tool returns the three highest-scoring chunks, after which the agent may issue another query or produce a final answer. This setup isolates the effect of agentic dense retrieval from both full-context prompting and LLM-based parallel document reading.
Retrieval operates over fixed-size chunks rather than over the entire paragraph or individual sentences, consistent with other chunk-based approaches. It use the same chunking strategy and training hyperparamemters as for \alg as described in \S\ref{app:training_setup}, but use a chunk size of $1{,}024$-token for inference.

Qwen3-Embedding-4B \citep{qwen3embedding}\footnote{\url{https://huggingface.co/Qwen/Qwen3-Embedding-4B}} is used as the retriever. Let $\bm{e}_u$ denote the normalized query embedding and $\bm{e}_i$ the normalized embedding of chunk $\bm{d}_i$. We rank chunks by
\[
    s(u,\bm{d}_i)=\bm{e}_u^\top\bm{e}_i,
\]
which is equivalent to cosine similarity after normalization, and return the top three. Each tool observation is a JSON object containing the original query and, for every retrieved chunk, its rank, similarity score, corpus index, and full text.
The agent runs in ReAct style. At each turn it generates at temperature $0$ with a maximum of $2{,}048$ new tokens. We allow at most $16$ turns. The system prompt explicitly encourages query reformulation and decomposition when the first retrieved set is incomplete, which permits multi-hop evidence to be collected over multiple retrieval rounds.

\begin{promptbox}{System prompt:}
\small\ttfamily
You are an intelligent helper responsible for answering complex multi-hop questions over a long document collection. You can use the `retrieve\_documents' tool to retrieve the three chunks most relevant to a search query.

\par\vspace{0.5\baselineskip}
\# How to work

\par\vspace{0.5\baselineskip}
1. Think about what evidence is needed, then call `retrieve\_documents' with a clear, focused natural-language query.

2. Read the retrieved chunks and decide whether they are sufficient. Evidence for a multi-hop answer may require multiple retrieval calls with different queries.

3. If results are partial or irrelevant, simplify, rephrase, or decompose the query instead of giving up.

4. When you can answer the question, put only the concise final answer inside \textcolor{Plum}{<answer>your answer</answer>}.

\par\vspace{0.5\baselineskip}
Do not claim that information is unavailable until you have tried multiple focused retrieval strategies.
\end{promptbox}

\begin{promptbox}{User prompt:}
\small\ttfamily
\#\# Question

\par\vspace{0.5\baselineskip}
\texttt{\{question\}}
\end{promptbox}

\subsection{MemAgent and ReMemR1}
\label{app:memagent_details}
We reproduce MemAgent \citep{yu2026memagent} and ReMemR1 \citep{shi2026look} from the authors' open-source repositories, adapting them only as needed to support Qwen3.5 \citep{qwen35blog} by upgrading the underlying VERL \citep{10.1145/3689031.3696075} dependency. We largely reuse the original training hyperparameters; the only intentional change is reducing the rollout group size from $16$ to $8$ (versus $5$ for \alg) to control training cost. Models are trained on $8$ NVIDIA H100 GPUs for $250$ steps, and a checkpoint is saved every 10 steps. Evaluation likewise follows the authors' released evaluation code. We refer readers to the original papers and code repositories for further implementation details.

\section{{P\scalebox{0.75}{AR}S\scalebox{0.75}{ER}\xspace}'s Implementation Details}
\label{app:training_details}

\subsection{Lead Agent Prompt}
\label{app:lead-prompt}
The system prompt contains tool call function usage and QA task instructions. We use the default tool call function template of Qwen3.5, which the models readily follow.

\begin{promptbox}{System prompt:}
\small\ttfamily
\# Tools

\par\vspace{0.5\baselineskip}
You have access to the following functions:

\par\vspace{0.5\baselineskip}
\textcolor{Plum}{\texttt{<tools>}}

\texttt{\{"type": "function", "function": \{}

\hspace*{1em}\texttt{"name": "query\_agents",}

\hspace*{1em}\texttt{"description": "Broadcast a query to all document agents in parallel. Each agent reads only its chunk and may return evidence and a partial answer. The tool response is a JSON string of aggregated findings.",}

\hspace*{1em}\texttt{"parameters": \{}

\hspace*{2em}\texttt{"type": "object", "properties": \{}

\hspace*{3em}\texttt{"query": \{"type": "string", "description": "Question or search query to send to every agent."\}}

\hspace*{2em}\texttt{\}, "required": ["query"]}

\hspace*{1em}\texttt{\}}

\texttt{\}}

\texttt{\}}

\textcolor{Plum}{\texttt{</tools>}}

\par\vspace{0.5\baselineskip}
If you choose to call a function ONLY reply in the following format with NO suffix:

\par\vspace{0.5\baselineskip}
\textcolor{Plum}{\texttt{<tool\_call>}}

\textcolor{Plum}{\texttt{<function=example\_function\_name>}}

\textcolor{Plum}{\texttt{<parameter=example\_parameter\_1>}}

value\_1

\textcolor{Plum}{\texttt{</parameter>}}

\textcolor{Plum}{\texttt{<parameter=example\_parameter\_2>}}

This is the value for the second parameter that can span\newline
multiple lines

\textcolor{Plum}{\texttt{</parameter>}}

\textcolor{Plum}{\texttt{</function>}}

\textcolor{Plum}{\texttt{</tool\_call>}}

\par\vspace{0.5\baselineskip}
\textcolor{Plum}{\texttt{<IMPORTANT>}}
Reminder:

- Function calls MUST follow the specified format: an inner \textcolor{Plum}{\texttt{<function=...></function>}} block must be nested within \textcolor{Plum}{\texttt{<tool\_call></tool\_call>}} XML tags.

- Required parameters MUST be specified.

- You may provide optional reasoning for your function call in natural language BEFORE the function call, but NOT after.

- If there is no function call available, answer the question like normal with your current knowledge and do not tell the user about function calls.

\textcolor{Plum}{\texttt{</IMPORTANT>}}

\par\vspace{0.5\baselineskip}
You are an intelligent helper responsible for answering complex multi-hop questions. Multiple agents exist, each with access to a different chunk of a long document. The evidence you need may be spread across chunks. You must use the \texttt{query\_agents} tool to retrieve information, then synthesize the final answer.

\par\vspace{0.5\baselineskip}
\textbf{How to work (multi-turn)}

\par\vspace{0.5\baselineskip}
1. After thinking, when you need evidence from the document, call the tool \texttt{query\_agents} with a clear natural-language \texttt{query}. The same query is broadcast to every agent; each searches only its chunk.

2. Read the tool response (JSON summarizing agent findings) and decide whether you need another query.

\hspace*{1em}-- \textbf{CRITICAL: DO NOT GIVE UP EASILY.} If a query returns no results, or only partial results, you MUST NOT immediately conclude ``Information not available.''

\hspace*{1em}-- Instead, you MUST simplify, rephrase, or break down your query into smaller parts and call \texttt{query\_agents} again.

\hspace*{1em}-- You MUST make multiple different attempts to query the agents before giving up. Only output the final answer when you are absolutely certain no more information can be found after exhausting multiple search strategies.

3. After thinking, when you can answer the question, write the final answer only inside \textcolor{Plum}{\texttt{<answer>your answer</answer>}}.
\end{promptbox}

\begin{promptbox}{User prompt:}
\small\ttfamily
\textbf{Question}

\textcolor{Plum}{\texttt{\{question\}}}
\end{promptbox}

\subsection{Subagent Prompt}
\label{app:subagent-prompt}
\begin{promptbox}{User prompt:}
\small\ttfamily
You are a precise document analysis assistant. Your task is to answer questions based ONLY on the information in your assigned document chunk.

\par\vspace{1\baselineskip}
\textbf{Core Principles}

1. \textbf{Strict Evidence-Based}: Answer ONLY if the chunk contains explicit information.

2. \textbf{Verbatim Extraction}: Evidence must be exact quotes, not paraphrases.

3. \textbf{Binary Output}: Either provide JSON with evidence, or output exactly \texttt{Unknown}.

\par\vspace{0.5\baselineskip}
\textbf{Response Protocol}
\par\vspace{0.5\baselineskip}
\textbf{If Evidence Exists}

\texttt{\textasciigrave\textasciigrave\textasciigrave{}json}

\texttt{\{"evidence": "Exact verbatim quote from the chunk that supports the answer. Must be copy-paste accurate.", "answer": "Concise answer based on the evidence"\}}

\texttt{\textasciigrave\textasciigrave\textasciigrave}

\par\vspace{0.5\baselineskip}
\textbf{If No Evidence}

Output exactly \texttt{Unknown} without any other text.
\par\vspace{1\baselineskip}
\textbf{Examples}
\par\vspace{0.5\baselineskip}
\textbf{Example 1:} Document Chunk: \texttt{<A chunk of text that contains the information about the invention of the telephone>}

Question: Who invented the telephone?

Response:

\texttt{\textasciigrave\textasciigrave\textasciigrave{}json}

\texttt{\{"evidence": "In 1876, Alexander Graham Bell was awarded the first US patent for the invention of the telephone.", "answer": "Alexander Graham Bell"\}}

\texttt{\textasciigrave\textasciigrave\textasciigrave}
\par\vspace{0.5\baselineskip}
\textbf{Example 2:} Document Chunk: \texttt{<A chunk of text that contains no information about the invention of the telephone>}

Question: Who invented the telephone?

Response: \texttt{Unknown}
\par\vspace{1\baselineskip}
\textbf{Your Task}
\par\vspace{0.5\baselineskip}
\textbf{Document Chunk:} \textcolor{Plum}{\texttt{\{chunk\_content\}}}
\par\vspace{0.5\baselineskip}
\textbf{Question:} \textcolor{Plum}{\texttt{\{question\}}}
\par\vspace{0.5\baselineskip}
Analyze the chunk against the question. Output ONLY the required format above, JSON or \texttt{Unknown}. Do NOT provide any explanations or additional text.
\end{promptbox}

\subsection{Overall Workflow}
Algorithm~\ref{alg:longmas} shows the overall workflow of \alg.
If a lead-agent generation contains neither a parseable \texttt{query\_agents} tool call nor an \textcolor{Plum}{\texttt{<answer>}} block, the environment does not terminate the trajectory. Instead, it appends \textsc{InvalidActionHint} as the observation and continues the ReAct loop, prompting the lead agent to retry with a well-formed action. The hint text is:
\begin{promptbox}{\textsc{InvalidActionHint}:}
\small\ttfamily
Invalid step: start with a short Thinking section (what you know, what you need, what you will do), then either call query\_agents via \textcolor{Plum}{<tool\_call>}\ldots\textcolor{Plum}{</tool\_call>} with \{"query": "..."\}, or give the final answer as \textcolor{Plum}{<answer>your answer</answer>}.
\end{promptbox}

\algdef{SE}[FOR]{PFor}{EndPFor}[1]{\textbf{parallel for} #1 \textbf{do}}{\textbf{end parallel for}}
\begin{algorithm}[ht]
\caption{\alg.}
\label{alg:longmas}
\algrenewcommand\algorithmiccomment[1]{\hfill$\triangleright$\ {\footnotesize #1}}
\begin{algorithmic}[1]
\Require Question $\bm{q}$, document chunks $\{\bm{d}_i\}_{i=1}^{T}$, maximum number of steps $K$
\Ensure Lead agent's reasoning trajectory, final answer
\State $\mathcal{H} \gets [\textsc{SystemPrompt}, \textsc{UserPrompt}(\bm{q})]$ \Comment{initialize the lead agent's context}
\For{$t = 1$ to $K$}
    \State $(\bm{z}_t, \mathcal{A}_t) \gets \textsc{LeadAgent}(\mathcal{H})$ \Comment{$\bm{z}_t$: thinking content; $\mathcal{A}_t$: action (subagent calling/answering)}
    \If{$\textsc{ExtractAnswer}(\mathcal{A}_t) \ne \varnothing$}
        \State $\mathcal{H} \gets \mathcal{H} \cup \{(\bm{z}_t, \mathcal{A}_t)\}$ \Comment{append current turn's generation to $\mathcal{H}$}
        \State \Return $(\mathcal{H}, \textsc{ExtractAnswer}(\mathcal{A}_t))$
    \ElsIf{$\textsc{ExtractQueries}(\mathcal{A}_t) \ne \varnothing$}
        \PFor{$u \in \textsc{ExtractQueries}(\mathcal{A}_t)$} \Comment{queries in the same turn run concurrently}
            \State \textbf{parallel for} $i = 1$ \textbf{to} $T$ \textbf{do} \Comment{Scatter: process all chunks concurrently}
                \State $o_i \gets \textsc{Subagent}(u, \bm{d}_i)$
            \State \textbf{end parallel for}
            \State $\mathcal{R}_u \gets \textsc{Gather}\big(\{o_i\}_{i=1}^{T}\big)$ \Comment{Gather: aggregate local findings}
        \EndPFor
        \State $\mathcal{H} \gets \mathcal{H} \cup \{(\bm{z}_t, \mathcal{A}_t, \{\mathcal{R}_u\}_{u \in \textsc{ExtractQueries}(\mathcal{A}_t)})\}$ \Comment{append current turn's generation and observations to $\mathcal{H}$}
    \Else
        \State $\mathcal{H} \gets \mathcal{H} \cup \{(\bm{z}_t, \mathcal{A}_t,\textsc{InvalidActionHint})\}$ \Comment{observation flags an invalid generation}
    \EndIf

\EndFor
\State \Return $(\mathcal{H}, \textsc{ExtractAnswer}(\mathcal{A}_t))$
\end{algorithmic}
\end{algorithm}

\subsection{Training Setup}
\label{app:training_setup}
\paragraph{Environment and deployment.}
We train \alg with VERL's Megatron backend. We adopt the fully asynchronous RL training architecture,\footnote{\url{https://verl.readthedocs.io/en/latest/advance/fully_async.html}} which decouples policy updating and rollout onto separate GPUs and allows both stages to run continuously. Beyond higher training throughput, this setup yields an additional practical benefit in our training: subagents continue serving rollout requests while the lead agent policy is being updated, rather than remaining idle during the update. We use six NVIDIA H100 GPUs for the trainer: two GPUs for policy updates and four GPUs for SGLang-based lead agent rollouts. We deploy the subagents with a separate SGLang Model Gateway\footnote{\url{https://docs.sglang.io/docs/advanced_features/sgl_model_gateway}} cluster on 10 additional NVIDIA H100 GPUs.

We optimize KV cache reuse for faster subagent inference. According to \S\ref{sec:workflow}, across all lead-agent turns, despite having different queries, each subagent is persistently assigned a fixed document chunk. Note that the subagent prompt is ordered as \emph{fixed instructions} $\rightarrow$ \emph{assigned document chunk} $\rightarrow$ \emph{current query}. Consequently, when the lead agent issues a new query in a later turn, requests sent to the same chunk differ only in the query prompt and subsequent text; the instructions and the chunk, which constitute most of the prompt, remain an identical prefix. During the first such request, SGLang computes the prefix's KV states and stores them in its Radix Cache. For subsequent requests with that prefix, the server can retrieve the cached KV states and prefill only the new query suffix, avoiding repeated computation over the long chunk.

This reuse requires sending a request to an instance that already holds the relevant cached prefix. We therefore configure the SGLang Router with the \texttt{cache\_aware} policy. For each incoming subagent request, the router compares its prompt prefix with the prefixes cached by its serving instances and preferentially routes the request to the instance with the longest match. The router thus preserves cache locality across lead agent turns, while the Radix Cache performs the KV reuse within the selected instance. This combination reduces redundant chunk-prefill computation and accelerates subagent inference.

\paragraph{Training hyperparameters.}
Table~\ref{tab:longmas_training_hparams} lists the training configuration. We optimize the lead agent with GRPO, use five rollouts per prompt, and set the learning rate to $1\times10^{-6}$. The lead agent operates for at most 9 turns; each subagent reads a 512-token chunk and generates at most 512 tokens per query.

\begin{longtable}{@{}p{0.23\textwidth}p{0.46\textwidth}p{0.25\textwidth}@{}}
\caption{Training hyperparameters for \alg.}\label{tab:longmas_training_hparams}\\
\toprule
\textbf{Category} & \textbf{Hyperparameter} & \textbf{Value} \\
\midrule
\endfirsthead
\toprule
\textbf{Category} & \textbf{Hyperparameter} & \textbf{Value} \\
\midrule
\endhead
\bottomrule
\endfoot
Actor optimization & Optimizer & Adam \\
& Learning rate; warm-up steps; schedule & $1\times10^{-6}$; $15$; constant \\
& Adam $(\beta_1,\beta_2)$; weight decay & $(0.9,0.999)$; $0.01$ \\
& Gradient clipping & $1.0$ \\
& PPO mini-batch size& $128$ \\
& PPO clip range; entropy coefficient & $[0.2,0.2]$; $0$ \\
\midrule
Megatron backend & Precision & bfloat16 \\
& Tensor / pipeline / context / expert parallelism & $1 / 1 / 1 / 1$ \\
& Parameter / gradient / optimizer offload & enabled / enabled / enabled \\
\midrule
Asynchronous training & Rollout engine; mode & SGLang; asynchronous \\
& Rollout GPUs; policy-update GPUs & $4$; $2$ \\
& Rollout GPU memory utilization & $0.8$ \\
& Staleness threshold; parameter-sync interval & $2$; $4$ steps \\
& Dynamic sampling & enabled \\
& Checkpoint interval & $10$ steps \\
\midrule
Lead agent & Rollouts per prompt & $5$ \\
& Sampling temperature; top-$p$; top-$k$ & $1.0$; $1.0$; $-1$ \\
& Maximum  turns & $9$ \\
& Maximum generation per turn & $2{,}048$ tokens \\
\midrule
Subagents & Temperature; output length & $0.7$; $512$ tokens \\
\end{longtable}
Training for 180 steps takes about 312 hours for the 4B model and 400 hours for the 9B model.

\paragraph{Chunking.}
For HotpotQA and 2WikiMultiHopQA, we first recover paragraph boundaries by splitting each packed context at its original ``\texttt{Document N:}'' delimiters and prefix each resulting unit with its paragraph index. Adjacent paragraphs are then greedily packed, in their original order, into chunks of at most $chunk\_size$ tokens according to the corresponding Qwen3.5 tokenizer. A paragraph that individually exceeds this limit is divided into contiguous token windows.

\subsection{Training Dynamics}
\begin{figure*}[p]
    \centering
    \includegraphics[
        width=\textwidth,
        height=0.98\textheight,
        keepaspectratio
    ]{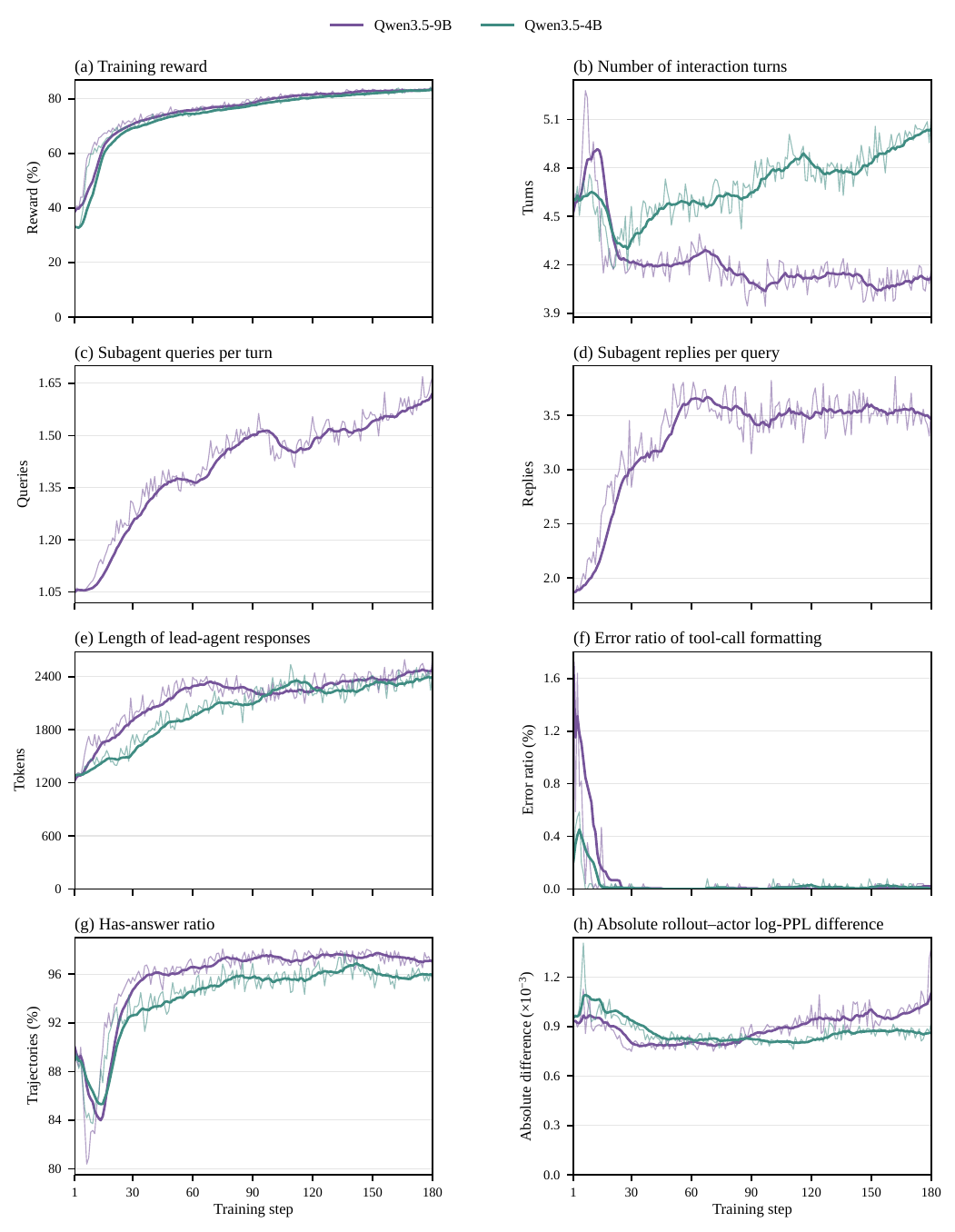}
    \caption{Training dynamics of the Qwen3.5-4B and Qwen3.5-9B lead agents over 180 reinforcement-learning steps. Thin curves show per-step measurements, while thick curves show 10-step moving averages. Logging for tool calls per turn and subagent replies per query was introduced after the 4B run had completed; consequently, these metrics are available only for the 9B run.}
    \label{fig:training_dynamics}
\end{figure*}

Figure~\ref{fig:training_dynamics} summarizes the training dynamics of the 4B and 9B lead agents. Comparing the averages over the first and last ten steps, the training reward increases from $44.9\%$ to $83.2\%$ for the 4B model and from $49.8\%$ to $83.3\%$ for the 9B model.
The two models reach similar rewards through different interaction dynamics. The mean number of turns increases from $4.65$ to $5.04$ for the 4B model, whereas it decreases from $4.85$ to $4.12$ for the 9B model. Meanwhile, the mean length of lead-agent responses grows from $1.36$K to $2.40$K tokens for 4B and from $1.47$K to $2.48$K tokens for 9B. For the 9B run, where the corresponding communication statistics were logged, subagent queries per turn increase steadily from $1.06$ to $1.62$. Notably, our reward neither penalizes the number of interaction turns nor explicitly encourages issuing multiple subagent queries in parallel. The increase therefore indicates an emergent strategy: the lead agent learns to identify queries without direct dependencies and place them in the same turn for parallel execution, rather than executing them sequentially across turns (see the example in Figure~\ref{fig:success_case_emma_woolf} and \ref{fig:success_case_inca}).

The lead agents also maintain reliable action formatting from the beginning of training: over the first ten steps, the tool-call format error ratio is only $0.22\%$ for 4B and $0.66\%$ for 9B, and it subsequently approaches zero. This reliability is achieved without a format reward because we use the backbone's native multi-turn tool-calling format, which the backbone has already been trained to follow. Finally, the absolute log-perplexity difference between the rollout and actor policies remains on the order of $10^{-3}$ throughout training (at most $1.41\times10^{-3}$ across both runs). Because rollout generation and policy optimization proceed concurrently, a trajectory may be generated by a rollout worker whose policy parameters lag behind the current actor by several updates. The consistently small discrepancy shows that this staleness causes only a minor shift in the token probabilities assigned to collected trajectories, thereby limiting the off-policy mismatch introduced by fully asynchronous training.

\subsection{Inference}
\label{app:inference_details}
Inference follows the same process as training. We serve the lead agent with a local SGLang offline engine on the trained checkpoint, and deploy subagents following the training setup described above.

The lead agent runs in thinking mode at temperature $0$ for up to $12$ turns and generates at most $2{,}048$ tokens per turn. Subagents use temperature $0.7$ and a $512$-token generation budget. Relative to training, evaluation uses larger chunks ($4{,}096$ vs.\ $512$ tokens) and a higher turn budget ($12$ vs.\ $9$). We intentionally adopt a smaller chunk size during training because finer chunking reduces prefill cost: as given in Equation~\eqref{eq:prefill_complexity}, the prefill complexity $\mathcal{O}(n^2/c)$ decreases as the number of chunks $c$ grows, which speeds up training. 

Following MemAgent \citep{yu2026memagent} and ReMemR1 \citep{shi2026look}, we report \texttt{Sub\_EM}: after standard answer normalization (lowercasing, removing articles and punctuation), a prediction is counted as correct if either the prediction or a ground-truth answer is a substring of the other.\footnote{Strictly speaking, \texttt{Sub\_EM} should be a unidirectional match in which the ground-truth answer is a substring of the model prediction; MemAgent's paper also describes the metric in this sense. However, the open-source evaluation code of MemAgent and ReMemR1 implements a bidirectional variant. We follow the latter definition.}

\section{Case Study}
\label{app:case_study}

\subsection{Success Case}
\label{app:success_cases}
In the example in Figure~\ref{fig:success_case_emma_woolf}, the lead agent decomposes the comparative question into two parallel \texttt{query\_agents} broadcasts that execute simultaneously, gathers birth-date evidence from different subagents, and synthesizes the final answer in a second reasoning step.
Figure~\ref{fig:success_case_cheese} shows a two-hop director lookup.The lead agent first recovers that \textit{I Want Someone to Eat Cheese With} was directed by Jeff Garlin, then queries his birth date.
Figure~\ref{fig:success_case_inca} shows a comparative multi-hop success trajectory. The lead agent broadcasts parallel director queries for \textit{Everything's Ducky} and \textit{Karthika}, then parallel death-date queries, and selects the film whose director died earlier.

Figure~\ref{fig:success_case_nungesser} illustrates how the lead agent resolves conflicting subagent responses. In the first round, different subagents return Fran\c{c}ois Coli alone, Charles Nungesser and Fran\c{c}ois Coli together, or Jean Metzinger. The last answer is triggered by the similarly named painting \textit{L'Oiseau bleu},(as mentioned in the subagent's evidence), rather than the target aircraft \textit{L'Oiseau Blanc}, and is therefore not adopted by the lead agent. Instead of committing to either of the remaining answers, the lead agent retains Nungesser and Coli as candidates and issues targeted parallel queries about both. The resulting evidence identifies Nungesser as the French ace pilot and adventurer, while describing Coli primarily as a pilot and navigator, allowing the lead agent to disambiguate the candidates and select Nungesser.

\subsection{Failure Case}
Figure~\ref{fig:failure_case_elene} shows a failure caused by a misleading subagent finding with the question
\textit{``Who is the husband of Princess Elene of Georgia?''} The displayed
trajectory contains two relevant findings. One correctly identifies the target
as the daughter of Heraclius~II of Georgia and the mother of Solomon~II of
Imereti. A second finding establishes that Solomon~II was born to Prince
Archil of Imereti and his wife Helene (an alternative name for Elene), the
daughter of Heraclius~II, which therefore supports the correct answer,
\textit{Prince Archil of Imereti}.
However, a subagent assigned to a chunk about Grand Duchess Elena Vladimirovna
of Russia returns the statement \textit{``Her husband was Prince
Nicholas of Greece and Denmark.''} Since the subagent only observes its local
chunk, it incorrectly treats the similarly named Elena in that chunk as the
Elene mentioned in the question. The lead agent subsequently accepts this
short, apparently direct answer and outputs \textit{Prince Nicholas of Greece
and Denmark}, despite the contradictory identity-grounded evidence from another
subagent.

This example exposes a failure pattern caused by \emph{context isolation
between the lead agent and subagents}. Each subagent receives only the lead
agent's query and a chunk, without the reasoning trajectory. When the query is underspecified or ambiguous, as in this case, the subagent may not recover the lead agent's current objective and can return an unintended conclusion based solely on its local chunk. Conversely, the lead agent has no access to the subagent's source reference and thus cannot directly verify whether the returned conclusion is grounded in a relevant reference. It may consequently over-trust an erroneous subagent conclusion. This behavior is occasional: in other cases, the lead agent issues additional queries that are more specific and reconciles evidence from multiple subagents before answering. The present failure occurs when that cross-validation process is not triggered or does not override the misleading local finding.

\subsection{Comparison against MemAgent on a Reverse-evidence Case}
\label{app:memagent_reverse}
Figure~\ref{fig:memagent_qd_sample} shows a question--document sample from the reverse-evidence setting.
The question is a comparative two-hop query: identify the directors of \textit{Everything's Ducky} and \textit{Karthika}, then select the film whose director died earlier.
The four supporting paragraphs are embedded in a 6{,}400-paragraph document.
Director death dates appear first (paragraphs 483 and 3911), while the film--director mappings appear later (paragraphs 5527 and 5631), i.e., in reverse logical order.

Figure~\ref{fig:memagent_memory} shows MemAgent's sequential memory updates on this sample.
When the agent encounters paragraph 483 (M.\ Krishnan Nair, died 2001) and later paragraph 3911 (Don Taylor, died 1998), it does not write either person's identity or death date into memory.
At those steps the two names have not yet been linked to the films in the question and are not written into the memory. Instead, the memory keeps recording directors of other films mentioned in the incoming chunks.
Only after paragraphs 5527 and 5631 does the memory record that \textit{Karthika} was directed by M.\ Krishnan Nair and \textit{Everything's Ducky} by Don Taylor.
By then the death dates have already been dropped.
The final memory therefore contains the two director names without their death dates, and MemAgent cannot complete the comparison. This trajectory shows that MemAgent's reasoning is heavily driven by the document: the order and surface content of incoming chunks determine what is written into memory.

By contrast, Figure~\ref{fig:success_case_inca} shows \alg on the same question.
\alg reasons from the question, issuing multi-round queries that execute in parallel over the full document, and is therefore completely insensitive to the order of evidence.

\clearpage
\clearpage

\clearpage

\definecolor{CaseQGreen}{HTML}{E8F5E9}
\definecolor{CaseQPurple}{HTML}{F3E5F5}
\definecolor{CaseQBlue}{HTML}{E3F2FD}
\definecolor{CaseQOrange}{HTML}{FFF3E0}
\def\caseqo#1#2{%
  \begin{tcolorbox}[
    colback=#1, colframe=#1, boxrule=0pt, arc=1.5pt,
    left=3pt, right=3pt, top=1.5pt, bottom=1.5pt,
    before skip=0pt, after skip=0pt,
    nobeforeafter,
  ]
  \ttfamily\scriptsize\raggedright #2%
  \end{tcolorbox}}
\def\casedoc#1{%
  \begin{tcolorbox}[
    colback=white, colframe=black, boxrule=0.55pt, arc=1.5pt,
    left=3pt, right=3pt, top=1.5pt, bottom=1.5pt,
    before skip=0pt, after skip=0pt,
    nobeforeafter,
  ]
  \scriptsize\raggedright #1%
  \end{tcolorbox}}
\newcommand{\casedots}{\par\vspace{5pt}\noindent\ldots\par\vspace{5pt}}
\tcbset{
  casestep/.style={
    enhanced, breakable,
    colback=white,
    colframe=white,
    boxrule=0pt,
    borderline={0.55pt}{0pt}{black!45,dash pattern=on 2.4pt off 1.6pt},
    arc=2.5pt, left=4pt, right=4pt, top=4pt, bottom=3pt,
    before skip=2pt, after skip=2pt,
    fontupper=\scriptsize,
    fonttitle=\bfseries\scriptsize, coltitle=black, colbacktitle=white,
    attach boxed title to top left={xshift=2.15em, yshift=-1.6mm},
    boxed title style={
      enhanced,
      boxrule=0pt,
      colframe=white,
      colback=white,
      frame hidden,
      overlay={},
      top=0pt, bottom=0pt, left=1pt, right=1pt,
    },
  },
}

\begingroup
\tcbset{casestep/.append style={unbreakable}}
\noindent
\begin{tcolorbox}[
  enhanced, breakable,
  colframe=black,
  colback=white,
  boxrule=0.85pt,
  arc=1pt,
  left=5pt,
  right=5pt,
  top=4pt,
  bottom=4pt,
  width=\linewidth,
]
\begin{tcolorbox}[casestep, title=Question]
{\footnotesize\textbf{User:}}\\[1pt]
Who was born earlier, Emma Bull or Virginia Woolf?
\end{tcolorbox}

\vspace{1pt}

\begin{tcolorbox}[casestep, title=Step 1]
{\footnotesize\textbf{Assistant:}}\\[1pt]
\textcolor{Plum}{\texttt{\textless think\textgreater}}
I need to find information about the birth dates of Emma Bull and Virginia Woolf to determine who was born earlier. Let me start by querying the agents to find information about both individuals.
\textcolor{Plum}{\texttt{\textless/think\textgreater}}

\vspace{2pt}
\caseqo{CaseQGreen}{\{"name": "query\_agents", "arguments": \{"query": "When was Emma Bull born?"\}\}}
\vspace{1pt}%
\caseqo{CaseQPurple}{\{"name": "query\_agents", "arguments": \{"query": "When was Virginia Woolf born?"\}\}}

\vspace{2pt}
\textbf{Observations:}\\[1pt]
\caseqo{CaseQGreen}{\{"agent\_129": \{"evidence": "Emma Bull (born December 13, 1954) is an American science fiction and fantasy author.", "answer": "December 13, 1954"\}\}}
\vspace{1pt}%
\caseqo{CaseQPurple}{\{"agent\_92": \{"evidence": "Adeline Virginia Woolf (n\'ee Stephen; 25 January 1882 -- 28 March 1941) was an English writer...", "answer": "25 January 1882"\}\}}
\end{tcolorbox}

\vspace{1pt}

\begin{tcolorbox}[casestep, title=Step 2]
{\footnotesize\textbf{Assistant:}}\\[1pt]
\textcolor{Plum}{\texttt{\textless think\textgreater}}
Great! I found the birth dates:\\
\hspace*{1em}-- Emma Bull: December 13, 1954\\
\hspace*{1em}-- Virginia Woolf: 25 January 1882\\
Comparing these dates, Virginia Woolf was born in 1882 and Emma Bull was born in 1954. Therefore, Virginia Woolf was born earlier than Emma Bull.

The answer is Virginia Woolf.
\textcolor{Plum}{\texttt{\textless/think\textgreater}}

\vspace{2pt}
Based on the information retrieved:\\
\hspace*{1em}-- Virginia Woolf was born on 25 January 1882\\
\hspace*{1em}-- Emma Bull was born on December 13, 1954\\
Virginia Woolf was born earlier than Emma Bull by approximately 72 years.

\vspace{2pt}
\textcolor{Plum}{\texttt{\textless answer\textgreater}}Adeline Virginia Woolf\textcolor{Plum}{\texttt{\textless/answer\textgreater}}
\end{tcolorbox}
\end{tcolorbox}
\captionof{figure}{Success case 1 (\alg-9B; HotpotQA, 6400-paragraph): the lead agent broadcasts two birth-date queries that execute simultaneously, gathers subagent findings, and synthesizes a comparative answer.}
\label{fig:success_case_emma_woolf}

\vspace{16pt}
\noindent
\begin{tcolorbox}[
  enhanced, breakable,
  colframe=black,
  colback=white,
  boxrule=0.85pt,
  arc=1pt,
  left=5pt,
  right=5pt,
  top=4pt,
  bottom=4pt,
  width=\linewidth,
]
\begin{tcolorbox}[casestep, title=Question]
{\footnotesize\textbf{User:}}\\[1pt]
When was the director of film I Want Someone To Eat Cheese With born?
\end{tcolorbox}

\vspace{1pt}

\begin{tcolorbox}[casestep, title=Step 1]
{\footnotesize\textbf{Assistant:}}\\[1pt]
\textcolor{Plum}{\texttt{\textless think\textgreater}}
I need to find information about the film ``I Want Someone To Eat Cheese With'' and then find out when its director was born.

Let me start by querying the agents to find information about this film.
\textcolor{Plum}{\texttt{\textless/think\textgreater}}

\vspace{2pt}
\caseqo{CaseQGreen}{\{"name": "query\_agents", "arguments": \{"query": "film I Want Someone To Eat Cheese With director"\}\}}

\vspace{2pt}
\textbf{Observations:}\\[1pt]
\caseqo{CaseQGreen}{\{"agent\_99": \{"evidence": "I Want Someone to Eat Cheese With is a 2006 American independent romantic comedy film written, produced, directed by, and...", "answer": "Jeff Garlin"\}\}}
\end{tcolorbox}

\vspace{1pt}

\begin{tcolorbox}[casestep, title=Step 2]
{\footnotesize\textbf{Assistant:}}\\[1pt]
\textcolor{Plum}{\texttt{\textless think\textgreater}}
Great! I found that the film ``I Want Someone To Eat Cheese With'' was directed by Jeff Garlin. Now I need to find out when Jeff Garlin was born. Let me query the agents for this information.
\textcolor{Plum}{\texttt{\textless/think\textgreater}}

\vspace{2pt}
\caseqo{CaseQGreen}{\{"name": "query\_agents", "arguments": \{"query": "Jeff Garlin birth date"\}\}}

\vspace{2pt}
\textbf{Observations:}\\[1pt]
\caseqo{CaseQGreen}{\{"agent\_130": \{"evidence": "Jeffrey Todd Garlin (born June 5, 1962) is an American comedian and actor.", "answer": "June 5, 1962"\}\}}
\end{tcolorbox}

\vspace{1pt}

\begin{tcolorbox}[casestep, title=Step 3]
{\footnotesize\textbf{Assistant:}}\\[1pt]
\textcolor{Plum}{\texttt{\textless think\textgreater}}
Perfect! I now have all the information I need:\\
\hspace*{1em}-- The film ``I Want Someone To Eat Cheese With'' was directed by Jeff Garlin\\
\hspace*{1em}-- Jeff Garlin was born on June 5, 1962

So the answer to the question ``When was the director of film I Want Someone To Eat Cheese With born?'' is June 5, 1962.
\textcolor{Plum}{\texttt{\textless/think\textgreater}}

\vspace{2pt}
Based on my research, I found that:

1. The film ``I Want Someone To Eat Cheese With'' (2006) was written, produced, directed by, and starring Jeff Garlin
2. Jeff Garlin (full name: Jeffrey Todd Garlin) was born on June 5, 1962

\vspace{2pt}
\textcolor{Plum}{\texttt{\textless answer\textgreater}}June 5, 1962\textcolor{Plum}{\texttt{\textless/answer\textgreater}}
\end{tcolorbox}
\end{tcolorbox}

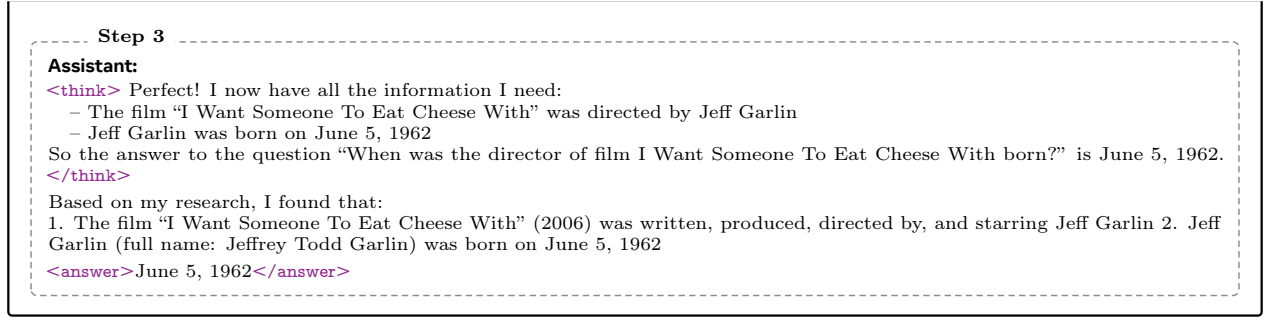
\captionof{figure}{Success case 2 (\alg-9B; 2WikiMultiHopQA, 6400-paragraph): the lead agent first retrieves the film's director, then queries the birth date and synthesizes the answer.}
\label{fig:success_case_cheese}

\vspace{16pt}
\noindent
\begin{tcolorbox}[
  enhanced, breakable,
  colframe=black,
  colback=white,
  boxrule=0.85pt,
  arc=1pt,
  left=5pt,
  right=5pt,
  top=4pt,
  bottom=4pt,
  width=\linewidth,
]
\begin{tcolorbox}[casestep, title=Question]
{\footnotesize\textbf{User:}}\\[1pt]
Which film has the director who died earlier, Everything's Ducky or Karthika (Film)?
\end{tcolorbox}

\vspace{1pt}

\begin{tcolorbox}[casestep, title=Step 1]
{\footnotesize\textbf{Assistant:}}\\[1pt]
\textcolor{Plum}{\texttt{\textless think\textgreater}}
This is a multi-hop question that requires me to find information about two films: ``Everything's Ducky'' and ``Karthika''. I need to find out who directed each film, and then determine which director died earlier.

Let me start by querying the agents to find information about these films and their directors.
\textcolor{Plum}{\texttt{\textless/think\textgreater}}

\vspace{2pt}
\caseqo{CaseQGreen}{\{"name": "query\_agents", "arguments": \{"query": "Everything's Ducky film director"\}\}}
\vspace{1pt}%
\caseqo{CaseQPurple}{\{"name": "query\_agents", "arguments": \{"query": "Karthika film director"\}\}}

\vspace{2pt}
\textbf{Observations:}\\[1pt]
\caseqo{CaseQGreen}{\{"agent\_15": \{"evidence": "Everything's Ducky is a 1961 comedy film directed by Don Taylor", "answer": "Don Taylor"\}\}}
\vspace{1pt}%
\caseqo{CaseQPurple}{\{"agent\_121": \{"evidence": "Karthika is a 1968 Indian Malayalam film, directed by M. Krishnan Nair", "answer": "M. Krishnan Nair"\}\}}
\end{tcolorbox}

\vspace{1pt}

\begin{tcolorbox}[casestep, title=Step 2]
{\footnotesize\textbf{Assistant:}}\\[1pt]
\textcolor{Plum}{\texttt{\textless think\textgreater}}
Great! Now I have the information that:\\
\hspace*{1em}-- ``Everything's Ducky'' is directed by Don Taylor\\
\hspace*{1em}-- ``Karthika (Film)'' is directed by M. Krishnan Nair

Now I need to find out which of these two directors died earlier. Let me query the agents for information about their death dates.
\textcolor{Plum}{\texttt{\textless/think\textgreater}}

\vspace{2pt}
\caseqo{CaseQGreen}{\{"name": "query\_agents", "arguments": \{"query": "Don Taylor director death date"\}\}}
\vspace{1pt}%
\caseqo{CaseQPurple}{\{"name": "query\_agents", "arguments": \{"query": "M. Krishnan Nair death date"\}\}}

\vspace{2pt}
\textbf{Observations:}\\[1pt]
\caseqo{CaseQGreen}{\{"agent\_172": \{"evidence": "Donald Richie Taylor (December 13, 1920 -- December 29, 1998) was an American actor and film director.", "answer": "December 29, 1998"\}\}}
\vspace{1pt}%
\caseqo{CaseQPurple}{\{"agent\_175": \{"evidence": "M. Krishnan Nair (2 November 1926 -- 10 May 2001 in Thiruvananthapuram) was an Indian film director of Malayalam films.", "answer": "10 May 2001"\}\}}
\end{tcolorbox}

\vspace{1pt}

\begin{tcolorbox}[casestep, title=Step 3]
{\footnotesize\textbf{Assistant:}}\\[1pt]
\textcolor{Plum}{\texttt{\textless think\textgreater}}
Perfect! Now I have the death dates for both directors:

\hspace*{1em}-- Don Taylor: December 29, 1998\\
\hspace*{1em}-- M. Krishnan Nair: 10 May 2001

Don Taylor died earlier (December 29, 1998) than M. Krishnan Nair (10 May 2001). Therefore, the film "Everything's Ducky" has the director who died earlier.
\textcolor{Plum}{\texttt{\textless/think\textgreater}}

\vspace{2pt}

\textcolor{Plum}{\texttt{\textless answer\textgreater}}Everything's Ducky\textcolor{Plum}{\texttt{\textless/answer\textgreater}}
\end{tcolorbox}
\end{tcolorbox}

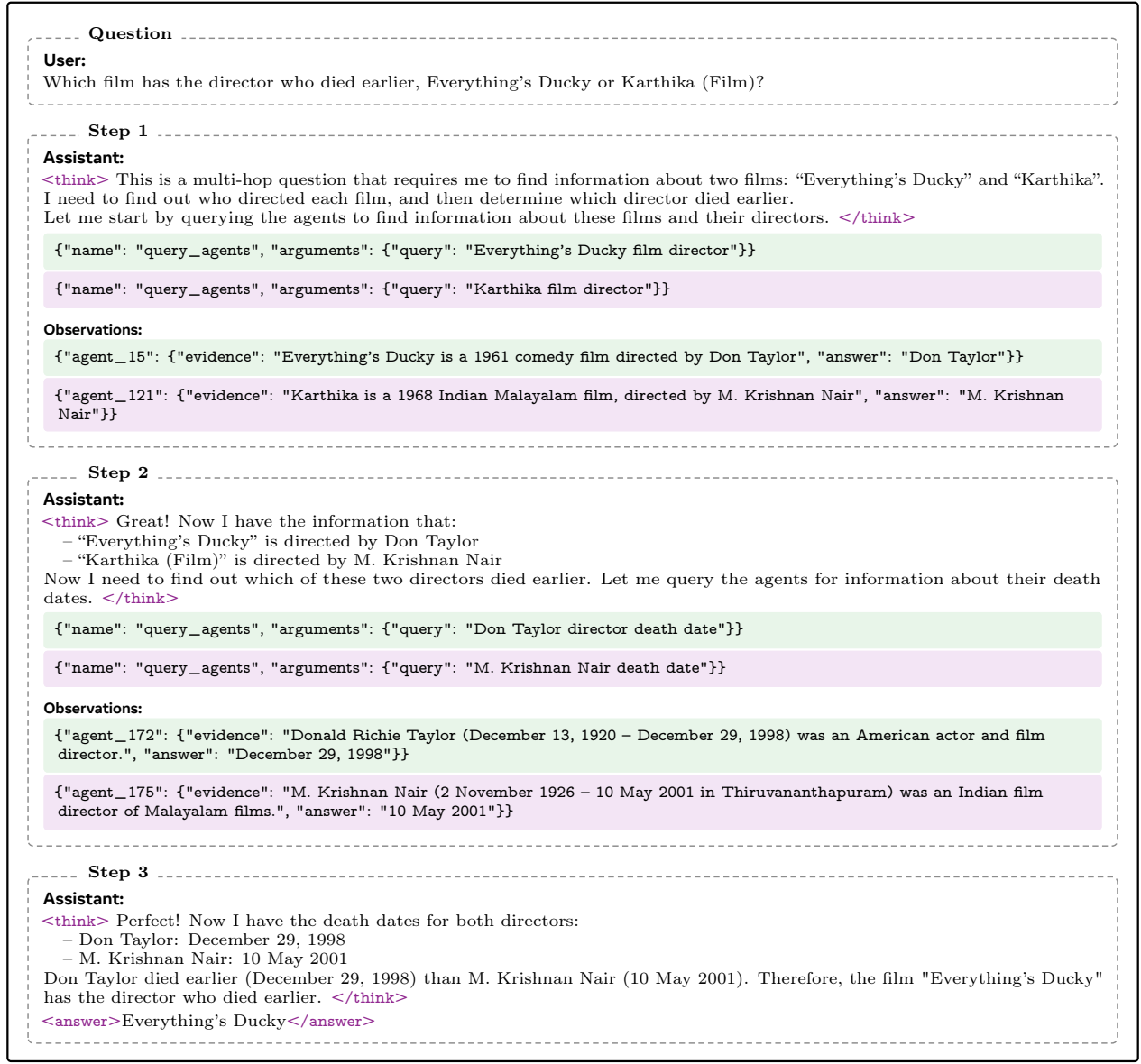
\captionof{figure}{Success case 3 (\alg-4B; 2WikiMultiHopQA, 6400-paragraph): the lead agent broadcasts parallel director queries, then parallel death-date queries, and selects the film whose director died earlier.}
\label{fig:success_case_inca}
\endgroup

\begin{figure}[p]
\centering
\footnotesize
\begin{tcolorbox}[
  enhanced,
  colframe=black,
  colback=white,
  boxrule=0.85pt,
  arc=1pt,
  left=5pt,
  right=5pt,
  top=4pt,
  bottom=4pt,
  width=\linewidth,
]
\begin{tcolorbox}[casestep, title=Question]
{\footnotesize\textbf{User:}}\\[1pt]
Which  French ace pilot and adventurer fly L'Oiseau Blanc?
\end{tcolorbox}

\vspace{1pt}

\begin{tcolorbox}[casestep, title=Step 1]
{\footnotesize\textbf{Assistant:}}\\[1pt]
\textcolor{Plum}{\texttt{\textless think\textgreater}}
I need to find information about a French ace pilot and adventurer who flew ``L'Oiseau Blanc''. This is a multi-hop question that requires me to:

1. First, find information about L'Oiseau Blanc and who flew it
2. Then identify if that person is a French ace pilot and adventurer

Let me start by querying the agents to find information about L'Oiseau Blanc.
\textcolor{Plum}{\texttt{\textless/think\textgreater}}

\vspace{2pt}
\caseqo{CaseQGreen}{\{"name": "query\_agents", "arguments": \{"query": "L'Oiseau Blanc who flew it"\}\}}

\vspace{2pt}
\textbf{Observations:}\\[1pt]
\caseqo{CaseQGreen}{\{"agent\_20": \{"evidence": "François Coli (June 5, 1881 -- presumably on or after May 8, 1927) was a French pilot and navigator best known as the one-eyed...", "answer": "François Coli"\}\} \\
\{"agent\_111": \{"evidence": "L'Oiseau bleu (also known as The Blue Bird and Der Blaue Vogel) is a large oil painting created in 1912--1913 by the French...", "answer": "Jean Metzinger"\}\} \\
\{"agent\_146": \{"evidence": "Only two examples of the type were built, with the first PL.8-01 named ''L'Oiseau Blanc'' (The White Bird), that gained fame as...", "answer": "Charles Nungesser and François Coli"\}\} \\
\{"agent\_167": \{"evidence": "The aircraft was flown by French World War I aviation heroes, Charles Nungesser and François Coli.", "answer": "Charles Nungesser and François Coli"\}\}}
\end{tcolorbox}

\vspace{1pt}

\begin{tcolorbox}[casestep, title=Step 2]
{\footnotesize\textbf{Assistant:}}\\[1pt]
\textcolor{Plum}{\texttt{\textless think\textgreater}}
Good, I found that L'Oiseau Blanc was flown by Charles Nungesser and François Coli. Now I need to find out which of these two is described as a ``French ace pilot and adventurer''. Let me search for information about both of them to see which one fits this description.

Let me search for information about Charles Nungesser first.
\textcolor{Plum}{\texttt{\textless/think\textgreater}}

\vspace{2pt}
\caseqo{CaseQGreen}{\{"name": "query\_agents", "arguments": \{"query": "Charles Nungesser French ace pilot adventurer"\}\}}
\vspace{1pt}%
\caseqo{CaseQPurple}{\{"name": "query\_agents", "arguments": \{"query": "François Coli French ace pilot adventurer"\}\}}

\vspace{2pt}
\textbf{Observations:}\\[1pt]
\caseqo{CaseQGreen}{\{"agent\_20": \{"evidence": "François Coli (June 5, 1881 -- presumably on or after May 8, 1927) was a French pilot and navigator best known as the one-eyed...", "answer": "François Coli"\}\} \\
\{"agent\_80": \{"evidence": "On the Paris side of the Atlantic their competitors were another World War One French ace, Charles Nungesser, and his navigator...", "answer": "The document identifies Charles Nungesser as a World War One French ace who was a competitor for the Orteig Prize alongside Charles Lindbergh in an attempt to fly the Atlantic Ocean."\}\} \\
\{"agent\_113": \{"evidence": "Charles Eugène Jules Marie Nungesser, MC (15 March 1892 -- presumably on or after 8 May 1927) was a French ace pilot and...", "answer": "Charles Eugène Jules Marie Nungesser, MC"\}\} \\
\{"agent\_146": \{"evidence": "Levasseur built the aircraft in 1927, specifically for pilots Charles Nungesser and François Coli for a transatlantic attempt to...", "answer": "Charles Nungesser was a French pilot who co-piloted a Levasseur PL.8 aircraft for a transatlantic attempt to win the Orteig Prize."\}\} \\
\{"agent\_167": \{"evidence": "L'Oiseau Blanc (commonly known in the English-speaking world as The White Bird) was a French Levasseur PL.8 biplane that...", "answer": "Charles Nungesser was a French World War I aviation hero who flew the aircraft L'Oiseau Blanc in an attempt to make the first non-stop transatlantic flight."\}\}}
\vspace{1pt}%
\caseqo{CaseQPurple}{\{"agent\_20": \{"evidence": "François Coli (June 5, 1881 -- presumably on or after May 8, 1927) was a French pilot and navigator best known as the one-eyed...", "answer": "François Coli was a French pilot and navigator, best known as the flying partner of Charles Nungesser in their attempt to fly the Atlantic Ocean."\}\} \\
\{"agent\_80": \{"evidence": "On the Paris side of the Atlantic their competitors were another World War One French ace, Charles Nungesser, and his navigator...", "answer": "No, François Coli was a navigator, not an ace pilot adventurer."\}\} \\
\{"agent\_146": \{"evidence": "Levasseur built the aircraft in 1927, specifically for pilots Charles Nungesser and François Coli for a transatlantic attempt to...", "answer": "François Coli was a French pilot who flew a Levasseur PL.8 aircraft in a transatlantic attempt to win the Orteig Prize."\}\} \\
\{"agent\_167": \{"evidence": "The aircraft was flown by French World War I aviation heroes, Charles Nungesser and François Coli.", "answer": "François Coli was a French World War I aviation hero."\}\}}
\end{tcolorbox}

\vspace{1pt}

\begin{tcolorbox}[casestep, title=Step 3]
{\footnotesize\textbf{Assistant:}}\\[1pt]
\textcolor{Plum}{\texttt{\textless think\textgreater}}
Perfect! I found the answer. According to agent\_113, ``Charles Eugène Jules Marie Nungesser, MC (15 March 1892 -- presumably on or after 8 May 1927) was a French ace pilot and adventurer, best remembered as a rival of Charles Lindbergh.''

This matches exactly with the question asking for ``Which French ace pilot and adventurer fly L'Oiseau Blanc''.

So the answer is Charles Nungesser (specifically Charles Eugène Jules Marie Nungesser).
\textcolor{Plum}{\texttt{\textless/think\textgreater}}

\vspace{2pt}
Based on my research, I found that Charles Nungesser (Charles Eugène Jules Marie Nungesser) was the French ace pilot and adventurer who flew L'Oiseau Blanc.
According to the evidence, Charles Nungesser was ``a French ace pilot and adventurer, best remembered as a rival of Charles Lindbergh.'' He flew the L'Oiseau Blanc aircraft along with his navigator François Coli in their attempt to make the first non-stop transatlantic flight from Paris to New York City in 1927.

\vspace{2pt}
\textcolor{Plum}{\texttt{\textless answer\textgreater}}Charles Eugène Jules Marie Nungesser\textcolor{Plum}{\texttt{\textless/answer\textgreater}}
\end{tcolorbox}
\end{tcolorbox}
\caption{Success case 4 (\alg-9B; HotpotQA, 6400-paragraph).}
\label{fig:success_case_nungesser}
\end{figure}

\clearpage
\noindent
\begin{tcolorbox}[
  enhanced,
  colframe=black,
  colback=white,
  boxrule=0.85pt,
  arc=1pt,
  left=5pt,
  right=5pt,
  top=4pt,
  bottom=4pt,
  width=\linewidth,
]
\begin{tcolorbox}[casestep, title=Question]
{\footnotesize\textbf{User:}}\\[1pt]
Who is the husband of Princess Elene Of Georgia?
\end{tcolorbox}

\vspace{1pt}

\begin{tcolorbox}[casestep, title=Step 1]
{\footnotesize\textbf{Assistant:}}\\[1pt]
\textcolor{Plum}{\texttt{\textless think\textgreater}}
This is a multi-hop question that requires me to find information about Princess Elene Of Georgia and her husband. I need to query the agents to find information about Princess Elene Of Georgia and then find information about her husband.

Let me start by querying the agents with a question about Princess Elene Of Georgia.
\textcolor{Plum}{\texttt{\textless/think\textgreater}}

\vspace{2pt}
\caseqo{CaseQGreen}{\{"name": "query\_agents", "arguments": \{"query": "Princess Elene Of Georgia"\}\}}

\vspace{2pt}
\textbf{Observations:}\\[1pt]
\caseqo{CaseQGreen}{\{"agent\_48": \{"evidence": "Elene (1753 -- 17 June 1786) was a Georgian princess royal (``batonishvili''), a daughter of Heraclius II, King of Kartli and Kakheti. She was the mother of Solomon II of Imereti, the last king to have reigned in the Georgian polities.", "answer": "Princess Elene of Georgia was a Georgian princess royal, daughter of Heraclius II, and mother of Solomon II of Imereti."\}\}}
\end{tcolorbox}

\vspace{1pt}

\begin{tcolorbox}[casestep, title=Step 2]
{\footnotesize\textbf{Assistant:}}\\[1pt]
\textcolor{Plum}{\texttt{\textless think\textgreater}}
I found information about Princess Elene of Georgia, but this doesn't mention her husband. The information shows she was a daughter of Heraclius II and mother of Solomon II of Imereti, but doesn't mention who her husband was.

I need to query again to find information about her husband. Let me try a different query focused on her husband or spouse.
\textcolor{Plum}{\texttt{\textless/think\textgreater}}

\vspace{2pt}
\caseqo{CaseQGreen}{\{"name": "query\_agents", "arguments": \{"query": "husband of Princess Elene Of Georgia"\}\}}

\vspace{2pt}
\textbf{Observations:}\\[1pt]
\caseqo{CaseQGreen}{\{"agent\_132": \{"evidence": "Her husband was Prince Nicholas of Greece and Denmark", "answer": "Prince Nicholas of Greece and Denmark"\}\} \\
\{"agent\_152": \{"evidence": "Solomon II (1772 – February 7, 1815), of the Bagrationi Dynasty, was the last King of Imereti (western Georgia) from 1789 to 1790 and from 1792 until his deposition by the Imperial Russian government in 1810. He was born as David to Prince Archil of Imereti, brother of King Solomon I of Imereti, by his wife Helene, daughter of King Heraclius II of Georgia.", "answer": "Solomon II"\}\}}
\end{tcolorbox}

\vspace{1pt}

\begin{tcolorbox}[casestep, title=Step 3]
{\footnotesize\textbf{Assistant:}}\\[1pt]
\textcolor{Plum}{\texttt{\textless think\textgreater}}
Great! I found the answer. The second query returned information from agent\_132 that states ``Her husband was Prince Nicholas of Greece and Denmark''. This directly answers the question ``Who is the husband of Princess Elene Of Georgia?''

The answer is Prince Nicholas of Greece and Denmark.
\textcolor{Plum}{\texttt{\textless/think\textgreater}}

\vspace{2pt}
\textcolor{Plum}{\texttt{\textless answer\textgreater}}Prince Nicholas of Greece and Denmark\textcolor{Plum}{\texttt{\textless/answer\textgreater}}
\end{tcolorbox}
\end{tcolorbox}
\captionof{figure}{Failure case (\alg-4B; 2WikiMultiHopQA, 6400-paragraph): a misleading local finding (\texttt{agent\_132}) overrides identity-grounded evidence (\texttt{agent\_152}) that supports the correct husband, Prince Archil of Imereti.}
\label{fig:failure_case_elene}

\vspace{16pt}
\noindent
\begin{tcolorbox}[
  enhanced,
  colframe=black,
  colback=white,
  boxrule=0.85pt,
  arc=1pt,
  left=5pt,
  right=5pt,
  top=4pt,
  bottom=4pt,
  width=\linewidth,
]
\begin{tcolorbox}[casestep, title=Question]
{\footnotesize}Which film has the director who died earlier, Everything's Ducky or Karthika (Film)?
\end{tcolorbox}

\vspace{1pt}

\begin{tcolorbox}[casestep, title=Document]
{\footnotesize Only evidence paragraphs are shown; the full document contains 6{,}400 paragraphs.}\\[2pt]
\ldots\\[2pt]
\casedoc{\textbf{Paragraph 483} \textit{M. Krishnan Nair (director)}\\
M. Krishnan Nair (2 November 1926 -- 10 May 2001 in Thiruvananthapuram) was an Indian film director of Malayalam films. He directed over 100 films. He also directed 18 Tamil movies including four films starring M. G. Ramachandran and two Telugu movies\ldots}
\vspace{2pt}
\ldots\\[2pt]
\casedoc{\textbf{Paragraph 3911} \textit{Don Taylor (American actor and director)}\\
Donald Richie Taylor (December 13, 1920 -- December 29, 1998) was an American actor and film director. He co-starred in 1940s and 1950s classics, including the 1948 film noir ``The Naked City'', ``Battleground'', ``Father of the Bride'', ``Father's Little Dividend'' and ``Stalag 17''. He later turned to directing films such as ``Escape from the Planet of the Apes'' (1971), ``Tom Sawyer'' (1973).}
\vspace{2pt}
\ldots\\[2pt]
\casedoc{\textbf{Paragraph 5527} \textit{Karthika (film)}\\
Karthika is a 1968 Indian Malayalam film, directed by M. Krishnan Nair and produced by V. M. Sreenivasan and A. R. Divakar. The film stars Sathyan, Sharada, Adoor Bhasi, Ummar and Manavalan Joseph in the lead roles. The film had musical score by M. S. Baburaj.}
\vspace{2pt}
\ldots\\[2pt]
\casedoc{\textbf{Paragraph 5631} \textit{Everything's Ducky}\\
Everything's Ducky is a 1961 comedy film directed by Don Taylor and written by Benedict Freedman and John Fenton Murray. The film stars Mickey Rooney, Buddy Hackett, Jackie Cooper, Joanie Sommers, Roland Winters and Elizabeth MacRae. The film was released on December 20, 1961, by Columbia Pictures.}
\vspace{2pt}
{\footnotesize\textit{\ldots (a total of 6{,}400 paragraphs)}}
\end{tcolorbox}
\end{tcolorbox}

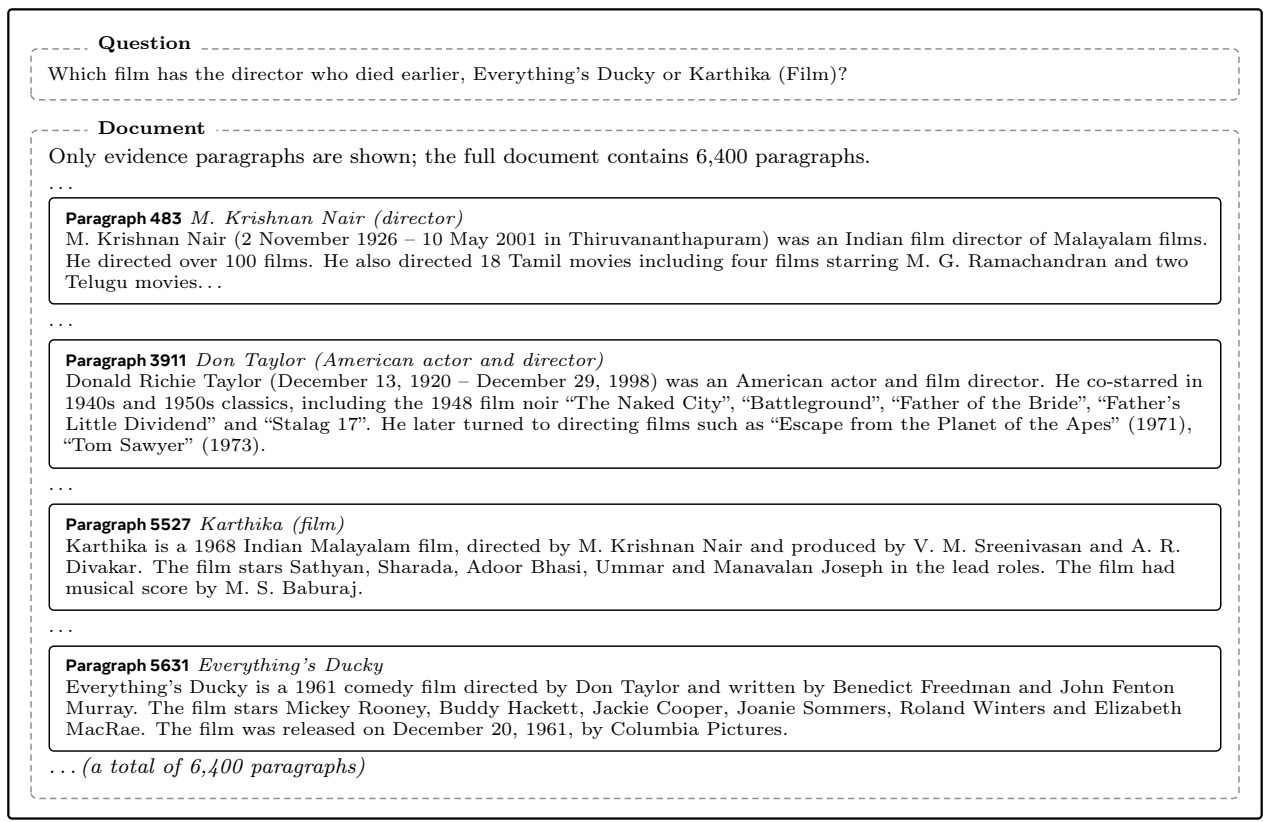
\captionof{figure}{Reverse-evidence question--document sample constructed from 2WikiMultiHopQA: death-date paragraphs (483, 3911) precede the film--director mappings (5527, 5631) in a 6{,}400-paragraph document. The correct answer is \textit{Everything's Ducky}.}
\label{fig:memagent_qd_sample}

\vspace{16pt}
\begingroup
\tcbset{casestep/.append style={unbreakable}}
\noindent
\begin{tcolorbox}[
  enhanced, breakable,
  colframe=black,
  colback=white,
  boxrule=0.85pt,
  arc=1pt,
  left=5pt,
  right=5pt,
  top=4pt,
  bottom=4pt,
  width=\linewidth,
]
{\footnotesize\textbf{MemAgent:}}
\casedots
\begin{tcolorbox}[casestep, title=Step 12 (after Paragraph 483)]
{\footnotesize\textbf{Memory:}}\\[1pt]
\hspace*{1em}-- \textbf{Problem:} Which film has the director who died earlier, Everything's Ducky or Karthika (Film)?\\
\hspace*{1em}-- Paragraph 11 states that Kim Jee-woon (born July 6, 1964) is a South Korean film director and screenwriter.\\
\hspace*{1em}-- Paragraph 7 states that the film \textit{Dharmakshetre Kurukshetre} is a 1975 Indian Malayalam film, directed and produced by Kunchacko.\\
\hspace*{1em}-- Paragraph 14 states that Wallace Fox (March 9, 1895 -- June 30, 1958) was an American film director. He directed 84 films between 1927 and 1953. He was born in Purcell, Oklahoma, and died in Hollywood, California.\\
\hspace*{1em}-- Paragraph 87 states that Anna of Trebizond is a film directed by a character who died in 1342.\\
\hspace*{1em}-- Paragraph 88 states that the film \textit{A Woman of the Sea} is a 1926 film directed by Josef von Sternberg.\\
\hspace*{1em}-- Paragraph 90 states that Tom DiCillo (Thomas A.\ ``Tom'' DiCillo) is an American film director, screenwriter and cinematographer.\\
\hspace*{1em}-- Paragraph 91 states that Annemarie S\"{o}rensen (19 December 1913 -- December 1993) was a German singer and film actress. She played the female lead in the 1934 comedy ``Heinz in the Moon''. She was a Protestant, but due to what was described as her uncertain racial origin she faced pressure following the Nazi takeover. In 1935 she emigrated from Germany and eventually went to live in Britain where she lived for the remainder of her life, although her attempt to break into British films failed.\\
\hspace*{1em}-- Paragraph 92 states that the film \textit{The Courtship of Andy Hardy} is a 1942 film, part of the Andy Hardy series. It gave an early role to Donna Reed although Mickey Rooney had lobbied for his then-wife Ava Gardner to have her part. Within a few months of the film's release, she filed for divorce. This was the 12th entry in the long-running ``Andy Hardy'' series.\\[2pt]
{\footnotesize\textcolor{blue}{\textit{Paragraph 483 (M.\ Krishnan Nair, died 2001) is not written into memory.}}}
\end{tcolorbox}

\casedots

\begin{tcolorbox}[casestep, title=Step 98 (after Paragraph 3911)]
{\footnotesize\textbf{Memory:}}\\[1pt]
\hspace*{1em}-- \textbf{Problem:} Which film has the director who died earlier, Everything's Ducky or Karthika (Film)?\\
\hspace*{1em}-- Paragraph 11 states that Kim Jee-woon (born July 6, 1964) is a South Korean film director and screenwriter.\\
\hspace*{1em}-- Paragraph 7 states that the film \textit{Dharmakshetre Kurukshetre} is a 1975 Indian Malayalam film, directed and produced by Kunchacko.\\
\hspace*{1em}-- Paragraph 1977 states that Louis Marie Malle (30 October 1932 -- 23 November 1995) was a French film director, screenwriter and producer. His film ``Le Monde du silence'' won the Palme d'Or in 1956 and the Academy Award for Best Documentary in 1957, although he was not credited at the ceremony; the award was instead presented to the film's co-director Jacques Cousteau. Later in his career he was nominated multiple times for Academy Awards. Malle is also one of only four directors to have won the Golden Lion twice. Malle worked in both French cinema and Hollywood, and he produced both French and English language films. His most famous films include the crime film ``Ascenseur pour l'\'{e}chafaud'' (1958), the World War II drama ``Lacombe, Lucien'' (1974), the romantic crime film ``Atlantic City'' (1980), the comedy-drama ``My Dinner with Andre'' (1981), and the autobiographical film ``Au revoir les enfants'' (1987).\\
\hspace*{1em}-- Paragraph 1978 states that Claude Autant-Lara (5 August 1901 -- 5 February 2000) was a French film director and later Member of the European Parliament (MEP).\\
\hspace*{1em}-- Paragraph 1979 states that Terror Train is a 1980 Canadian slasher film directed by Roger Spottiswoode in his directorial debut and starring Jamie Lee Curtis, Ben Johnson, and Hart Bochner. Set aboard a moving train on New Year's Eve, the film follows a group of medical school students holding a costume party who are targeted by a killer who steals their costumes after murdering them to avoid being caught. It features supporting performances from Sandee Currie, Anthony Sherwood, and David Copperfield. The concept for the film was based on an idea by executive producer Daniel Grodnik, who sought to ``make Halloween on a train.'' A full-length script for the film was composed by T.\ Y.\ Drake, and production was initiated within four months. The film was shot in Montreal between late November and late December.\\
\hspace*{1em}-- Paragraph 2311 states that Russ Meyer (Russell Albion Meyer) was an American film director, producer, screenwriter, cinematographer, film editor, actor, and photographer. Meyer is known primarily for writing and directing a series of successful sexploitation films that featured campy humor, sly satire and large-breasted women, such as ``Faster, Pussycat! Kill! Kill!'' He was born March 21, 1922, and died September 18, 2004.\\
\hspace*{1em}-- Paragraph 2317 states that Helmut K\"{a}utner (born 25 March 1908 in D\"{u}sseldorf, Germany; died 20 April 1980 in Castellina in Chianti, Italy) was a German film director active mainly in the 1940s and 1950s.\\
\hspace*{1em}-- Paragraph 2765 states that the film \textit{The Man with a Cloak} is a 1951 American film noir crime-thriller-drama directed by Fletcher Markle and starring Joseph Cotten, Barbara Stanwy.\\
\hspace*{1em}-- Paragraph 3591 states that the film \textit{The Devil's Hairpin} is a 1957 feature film about car racing, filmed in Technicolor and VistaVision, written and directed by Cornel Wilde, who also stars.\\
\hspace*{1em}-- Paragraph 3605 states that the film \textit{Sunshine Sally} is a 1922 Australian silent film directed by Lawson Harris set in the Sydney suburb of Woolloomooloo. Most of the movie survives today.\\
\hspace*{1em}-- Paragraph 3706 states that the film \textit{Toen 't licht verdween} is a 1918 Dutch silent drama film directed by Maurits Binger.\\[2pt]
{\footnotesize\textcolor{blue}{\textit{Paragraph 3911 (Don Taylor, died 1998) is not written into memory. The memory continues to store directors of other films.}}}
\end{tcolorbox}

\casedots

\begin{tcolorbox}[casestep, title=Step 140 (after Paragraph 5527)]
{\footnotesize\textbf{Memory:}}\\[1pt]
\hspace*{1em}-- \textbf{Problem:} Which film has the director who died earlier, Everything's Ducky or Karthika (Film)?\\
\hspace*{1em}-- Paragraph 5527 states that the film \textit{Karthika} is a 1968 Indian Malayalam film, directed by \textbf{M.\ Krishnan Nair} and produced by V.\ M.\ Sreenivasan and A.\ R.\ Divakar. The film stars Sathyan, Sharada, Adoor Bhasi, Ummar and Manavalan Joseph in the lead roles. The film had musical score by M.\ S.\ Baburaj.\\
\hspace*{1em}-- Paragraph 11 states that Kim Jee-woon (born July 6, 1964) is a South Korean film director and screenwriter.\\
\hspace*{1em}-- Paragraph 7 states that the film \textit{Dharmakshetre Kurukshetre} is a 1975 Indian Malayalam film, directed and produced by Kunchacko.\\
\hspace*{1em}-- Paragraph 1977 states that Louis Marie Malle (30 October 1932 -- 23 November 1995) was a French film director, screenwriter and producer. His film ``Le Monde du silence'' won the Palme d'Or in 1956 and the Academy Award for Best Documentary in 1957, although he was not credited at the ceremony; the award was instead presented to the film's co-director Jacques Cousteau. Later in his career he was nominated multiple times for Academy Awards. Malle is also one of only four directors to have won the Golden Lion twice. Malle worked in both French cinema and Hollywood, and he produced both French and English language films. His most famous films include the crime film ``Ascenseur pour l'\'{e}chafaud'' (1958), the World War II drama ``Lacombe, Lucien'' (1974), the romantic crime film ``Atlantic City'' (1980), the comedy-drama ``My Dinner with Andre'' (1981), and the autobiographical film ``Au revoir les enfants'' (1987).\\
\hspace*{1em}-- Paragraph 1978 states that Claude Autant-Lara (5 August 1901 -- 5 February 2000) was a French film director and later Member of the European Parliament (MEP).\\
\hspace*{1em}-- Paragraph 1979 states that Terror Train is a 1980 Canadian slasher film directed by Roger Spottiswoode in his directorial debut and starring Jamie Lee Curtis, Ben Johnson, and Hart Bochner. Set aboard a moving train on New Year's Eve, the film follows a group of medical school students holding a costume party who are targeted by a killer who steals their costumes after murdering them to avoid being caught. It features supporting performances from Sandee Currie, Anthony Sherwood, and David Copperfield. The concept for the film was based on an idea by executive producer Daniel Grodnik, who sought to ``make Halloween on a train.'' A full-length script for the film was composed by T.\ Y.\ Drake, and production was initiated within four months. The film was shot in Montreal between late November and late December.\\
\hspace*{1em}-- Paragraph 2311 states that Russ Meyer (Russell Albion Meyer) was an American film director, producer, screenwriter, cinematographer, film editor, actor, and photographer. Meyer is known primarily for writing and directing a series of successful sexploitation films that featured campy humor, sly satire and large-breasted women, such as ``Faster, Pussycat! Kill! Kill!'' He was born March 21, 1922, and died September 18, 2004.\\
\hspace*{1em}-- Paragraph 2317 states that Helmut K\"{a}utner (born 25 March 1908 in D\"{u}sseldorf, Germany; died 20 April 1980 in Castellina in Chianti, Italy) was a German film director active mainly in the 1940s and 1950s.\\
\hspace*{1em}-- Paragraph 2765 states that the film \textit{The Man with a Cloak} is a 1951 American film noir crime-thriller-drama directed by Fletcher Markle and starring Joseph Cotten, Barbara Stanwy.\\
\hspace*{1em}-- Paragraph 3591 states that the film \textit{The Devil's Hairpin} is a 1957 feature film about car racing, filmed in Technicolor and VistaVision, written and directed by Cornel Wilde, who also stars.\\[2pt]
{\footnotesize\textcolor{blue}{\textit{The film--director mapping is stored, but M.\ Krishnan Nair's death date is still absent.}}}
\end{tcolorbox}

\casedots

\begin{tcolorbox}[casestep, title=Step 142 (after Paragraph 5631)]
{\footnotesize\textbf{Memory:}}\\[1pt]
\hspace*{1em}-- \textbf{Problem:} Which film has the director who died earlier, Everything's Ducky or Karthika (Film)?\\
\hspace*{1em}-- Paragraph 5631 states that the film \textit{Everything's Ducky} is a 1961 comedy film directed by \textbf{Don Taylor} and written by Benedict Freedman and John Fenton Murray. The film stars Mickey Rooney, Buddy Hackett, Jackie Cooper, Joanie Sommers, Roland Winters and Elizabeth MacRae. The film was released on December 20, 1961, by Columbia Pictures.\\
\hspace*{1em}-- Paragraph 5527 states that the film \textit{Karthika} is a 1968 Indian Malayalam film, directed by \textbf{M.\ Krishnan Nair} and produced by V.\ M.\ Sreenivasan and A.\ R.\ Divakar. The film stars Sathyan, Sharada, Adoor Bhasi, Ummar and Manavalan Joseph in the lead roles. The film had musical score by M.\ S.\ Baburaj.\\
\hspace*{1em}-- Paragraph 11 states that Kim Jee-woon (born July 6, 1964) is a South Korean film director and screenwriter.\\
\hspace*{1em}-- Paragraph 7 states that the film \textit{Dharmakshetre Kurukshetre} is a 1975 Indian Malayalam film, directed and produced by Kunchacko.\\
\hspace*{1em}-- Paragraph 1977 states that Louis Marie Malle (30 October 1932 -- 23 November 1995) was a French film director, screenwriter and producer. His film ``Le Monde du silence'' won the Palme d'Or in 1956 and the Academy Award for Best Documentary in 1957, although he was not credited at the ceremony; the award was instead presented to the film's co-director Jacques Cousteau. Later in his career he was nominated multiple times for Academy Awards. Malle is also one of only four directors to have won the Golden Lion twice. Malle worked in both French cinema and Hollywood, and he produced both French and English language films. His most famous films include the crime film ``Ascenseur pour l'\'{e}chafaud'' (1958), the World War II drama ``Lacombe, Lucien'' (1974), the romantic crime film ``Atlantic City'' (1980), the comedy-drama ``My Dinner with Andre'' (1981), and the autobiographical film ``Au revoir les enfants'' (1987).\\
\hspace*{1em}-- Paragraph 1978 states that Claude Autant-Lara (5 August 1901 -- 5 February 2000) was a French film director and later Member of the European Parliament (MEP).\\
\hspace*{1em}-- Paragraph 1979 states that Terror Train is a 1980 Canadian slasher film directed by Roger Spottiswoode in his directorial debut and starring Jamie Lee Curtis, Ben Johnson, and Hart Bochner. Set aboard a moving train on New Year's Eve, the film follows a group of medical school students holding a costume party who are targeted by a killer who steals their costumes after murdering them to avoid being caught. It features supporting performances from Sandee Currie, Anthony Sherwood, and David Copperfield. The concept for the film was based on an idea by executive producer Daniel Grodnik, who sought to ``make Halloween on a train.'' A full-length script for the film was composed by T.\ Y.\ Drake, and production was initiated within four months. The film was shot in Montreal between late November and late December.\\
\hspace*{1em}-- Paragraph 2311 states that Russ Meyer (Russell Albion Meyer) was an American film director, producer, screenwriter, cinematographer, film editor, actor, and photographer. Meyer is known primarily for writing and directing a series of successful sexploitation films that featured campy humor, sly satire and large-breasted women, such as ``Faster, Pussycat! Kill! Kill!'' He was born March 21, 1922, and died September 18, 2004.\\
\hspace*{1em}-- Paragraph 2317 states that Helmut K\"{a}utner (born 25 March 1908 in D\"{u}sseldorf, Germany; died 20 April 1980 in Castellina in Chianti, Italy) was a German film director active mainly in the 1940s and 1950s.\\[2pt]
{\footnotesize\textcolor{blue}{\textit{Both target director names are stored, but neither death date is present.}}}
\end{tcolorbox}

\casedots

\begin{tcolorbox}[casestep, title=Step 161 (after all paragraphs)]
{\footnotesize\textbf{Memory:}}\\[1pt]
\hspace*{1em}-- \textbf{Problem:} Which film has the director who died earlier, Everything's Ducky or Karthika (Film)?\\
\hspace*{1em}-- Paragraph 5631 states that the film \textit{Everything's Ducky} is a 1961 comedy film directed by \textbf{Don Taylor} and written by Benedict Freedman and John Fenton Murray. The film stars Mickey Rooney, Buddy Hackett, Jackie Cooper, Joanie Sommers, Roland Winters and Elizabeth MacRae. The film was released on December 20, 1961, by Columbia Pictures.\\
\hspace*{1em}-- Paragraph 5527 states that the film \textit{Karthika} is a 1968 Indian Malayalam film, directed by \textbf{M.\ Krishnan Nair} and produced by V.\ M.\ Sreenivasan and A.\ R.\ Divakar. The film stars Sathyan, Sharada, Adoor Bhasi, Ummar and Manavalan Joseph in the lead roles. The film had musical score by M.\ S.\ Baburaj.\\
\hspace*{1em}-- Paragraph 1977 states that Louis Marie Malle (30 October 1932 -- 23 November 1995) was a French film director, screenwriter and producer. His film ``Le Monde du silence'' won the Palme d'Or in 1956 and the Academy Award for Best Documentary in 1957, although he was not credited at the ceremony; the award was instead presented to the film's co-director Jacques Cousteau. Later in his career he was nominated multiple times for Academy Awards. Malle is also one of only four directors to have won the Golden Lion twice. Malle worked in both French cinema and Hollywood, and he produced both French and English language films. His most famous films include the crime film ``Ascenseur pour l'\'{e}chafaud'' (1958), the World War II drama ``Lacombe, Lucien'' (1974), the romantic crime film ``Atlantic City'' (1980), the comedy-drama ``My Dinner with Andre'' (1981), and the autobiographical film ``Au revoir les enfants'' (1987).\\
\hspace*{1em}-- Paragraph 1978 states that Claude Autant-Lara (5 August 1901 -- 5 February 2000) was a French film director and later Member of the European Parliament (MEP).\\
\hspace*{1em}-- Paragraph 1979 states that Terror Train is a 1980 Canadian slasher film directed by Roger Spottiswoode in his directorial debut and starring Jamie Lee Curtis, Ben Johnson, and Hart Bochner. Set aboard a moving train on New Year's Eve, the film follows a group of medical school students holding a costume party who are targeted by a killer who steals their costumes after murdering them to avoid being caught. It features supporting performances from Sandee Currie, Anthony Sherwood, and David Copperfield. The concept for the film was based on an idea by executive producer Daniel Grodnik, who sought to ``make Halloween on a train.'' A full-length script for the film was composed by T.\ Y.\ Drake, and production was initiated within four months. The film was shot in Montreal between late November and late December.\\
\hspace*{1em}-- Paragraph 2311 states that Russ Meyer (Russell Albion Meyer) was an American film director, producer, screenwriter, cinematographer, film editor, actor, and photographer. Meyer is known primarily for writing and directing a series of successful sexploitation films that featured campy humor, sly satire and large-breasted women, such as ``Faster, Pussycat! Kill! Kill!'' He was born March 21, 1922, and died September 18, 2004.\\
\hspace*{1em}-- Paragraph 2317 states that Helmut K\"{a}utner (born 25 March 1908 in D\"{u}sseldorf, Germany; died 20 April 1980 in Castellina in Chianti, Italy) was a German film director active mainly in the 1940s and 1950s.\\
\hspace*{1em}-- Paragraph 6215 states that Rajkumar Kohli (born 14 September 1930 in Lahore, British India) is an Indian film director. He was famous for directing several popular Bollywood films in the 1970s like Lootera starring Dara Singh and Nishi, who married Rajkumar Koli and 1980s with ensemble cast such as ``Nagin'' (1976).\\[2pt]
{\footnotesize\textcolor{blue}{\textit{The final memory still has only the two director names, not their death dates.}}}
\end{tcolorbox}

\vspace{1pt}

\begin{tcolorbox}[casestep, title=Final answer]
Based on the provided memory, we can compare the directors of the films \textit{Everything's Ducky} and \textit{Karthika}:\\[1pt]
\hspace*{1em}1. The film \textit{Everything's Ducky} (1961) is directed by \textbf{Don Taylor}.\\
\hspace*{1em}2. The film \textit{Karthika} (1968) is directed by \textbf{M.\ Krishnan Nair}.\\
\hspace*{1em}3. The memory explicitly states that \textbf{Louis Marie Malle} (30 October 1932 -- 23 November 1995) was a French film director, screenwriter and producer. His film ``Le Monde du silence'' won the Palme d'Or in 1956 and the Academy Award for Best Documentary in 1957, although he was not credited at the ceremony; the award was instead presented to the film's co-director Jacques Cousteau. Later in his career he was nominated multiple times for Academy Awards. Malle is also one of only four directors to have won the Golden Lion twice. Malle worked in both French cinema and Hollywood, and he produced both French and English language films. His most famous films include the crime film ``Ascenseur pour l'\'{e}chafaud'' (1958), the World War II drama ``Lacombe, Lucien'' (1974), the romantic crime film ``Atlantic City'' (1980), the comedy-drama ``My Dinner with Andre'' (1981), and the autobiographical film ``Au revoir les enfants'' (1987).\\
\hspace*{1em}4. The memory also states that \textbf{Claude Autant-Lara} (5 August 1901 -- 5 February 2000) was a French film director and later Member of the European Parliament (MEP).\\
\hspace*{1em}5. The memory states that \textbf{Russ Meyer} (Russell Albion Meyer) was an American film director, producer, screenwriter, cinematographer, film editor, actor, and photographer. Meyer is known primarily for writing and directing a series of successful sexploitation films that featured campy humor, sly satire and large-breasted women, such as ``Faster, Pussycat! Kill! Kill!'' He was born March 21, 1922, and died September 18, 2004.\\
\hspace*{1em}6. The memory states that \textbf{Helmut K\"{a}utner} (born 25 March 1908 in D\"{u}sseldorf, Germany; died 20 April 1980 in Castellina in Chianti, Italy) was a German film director active mainly in the 1940s and 1950s.\\
\hspace*{1em}7. The memory states that \textbf{Rajkumar Kohli} (born 14 September 1930 in Lahore, British India) is an Indian film director. He was famous for directing several popular Bollywood films in the 1970s like Lootera starring Dara Singh and Nishi, who married Rajkumar Koli and 1980s with ensemble cast such as ``Nagin'' (1976), ``Jaani Dushman''.\\
\hspace*{1em}8. The memory states that \textbf{Rajkumar Kohli} was famous for directing several popular Bollywood films in the 1970s like Lootera starring Dara Singh and Nishi, who married Rajkumar Koli and 1980s with ensemble cast such as ``Nagin'' (1976), ``Jaani Dushman''.\\
\ldots\\[2pt]
{\footnotesize\textcolor{blue}{\textit{The two target death dates were never written, so the comparison cannot be completed.}}}
\end{tcolorbox}
\end{tcolorbox}
\captionof{figure}{MemAgent (4B)'s memory updates on the reverse-evidence sample. Death dates in paragraphs 483 and 3911 are skipped because their link to the question is not yet known; later updates keep only the two director names. Intermediate steps are omitted (``\ldots'').}
\label{fig:memagent_memory}
\endgroup

\end{document}